\documentclass[11pt]{article}

\PassOptionsToPackage{table}{xcolor}
\usepackage[preprint]{acl}
\usepackage{times}
\usepackage{latexsym}
\usepackage[T1]{fontenc}
\usepackage[utf8]{inputenc}
\usepackage{microtype}
\usepackage{inconsolata}
\usepackage{graphicx}
\usepackage{amsmath}
\usepackage{amssymb}
\usepackage{booktabs}
\usepackage{multirow}
\usepackage{algorithm}
\usepackage{algpseudocode}
\usepackage{enumitem}
\usepackage{subfig}
\usepackage[table]{xcolor}
\usepackage{framed}
\usepackage{textcomp}

\definecolor{TFFrameColor}{gray}{0.50}
\definecolor{TFTitleColor}{gray}{1.00}
\newenvironment{promptbox}[1]{%
  \nolinenumbers
  \small
  \setlength{\OuterFrameSep}{8pt}%
  \def\FrameCommand{\fboxsep7pt\fboxrule0.5pt
    \TitleBarFrame{\textbf{#1}}}%
  \let\FirstFrameCommand\FrameCommand
  \def\MidFrameCommand{\fboxsep7pt\fboxrule0.5pt
    \TitleBarFrame{\textbf{#1\ (continued)}}}%
  \let\LastFrameCommand\MidFrameCommand
  \MakeFramed{\advance\hsize-15pt \FrameRestore}%
  \setlength{\parindent}{0pt}%
  \setlength{\parskip}{4pt}%
}{\endMakeFramed}

\title{AlphaRLM: Aligning Language Models for Literature-Grounded Symbolic Factor Discovery}

\author{
  Jie Huang\textsuperscript{1}\thanks{Equal contribution.}, Yuecheng Hong\textsuperscript{1}\footnotemark[1], Yao Chen\textsuperscript{3}, Hao Zhong\textsuperscript{3} \\
  \textbf{Linfeng Xiao\textsuperscript{3}, Xudong Liu\textsuperscript{1}, Qiang Ye\textsuperscript{2}} \\
  \textbf{Hao Xia\textsuperscript{1}, Jianxuan Xin\textsuperscript{2}\thanks{Corresponding author.}} \\[0.5em]
  \textsuperscript{1}Harbin Institute of Technology \\
  \textsuperscript{2}University of Science and Technology of China \\
  \textsuperscript{3}Independent Researcher \\[0.4em]
  {\small\texttt{jhuang@ir.hit.edu.cn, jianxuan@ustc.edu.cn}}
}

\begin{document}
\maketitle
\begin{abstract}
Formulaic alpha mining seeks to discover executable expressions that capture predictive market signals in a vast, noisy search space.
Recent LLM-based approaches use prompting, iterative search, or seed-factor evolution, leaving direct factor construction through policy learning less explored.
To address this, we introduce AlphaRLM, a reinforcement-learning framework that trains LLMs to construct executable symbolic factors from financial literature without input factor expressions.
AlphaRLM incorporates Literature-Guided Operator Assembly (LEGO), which conditions factor construction on literature-derived ideas and executable operators, providing semantic priors for exploration.
It further aligns the generation policy with market feedback through executable-validity and grouped-return-separation rewards that capture non-monotonic structures across factor quantiles.
Extensive experiments on CSI 300, CSI 1000, and S\&P 500 show that AlphaRLM outperforms strong factor-mining and LLM-based baselines in annualized excess return.
Notably, the 8B model achieves strong cross-market performance, highlighting the promise of literature-conditioned, market-aligned factor generation.
\end{abstract}

\section{Introduction}

\begin{figure*}[t]
    \centering
    \includegraphics[width=0.92\textwidth]{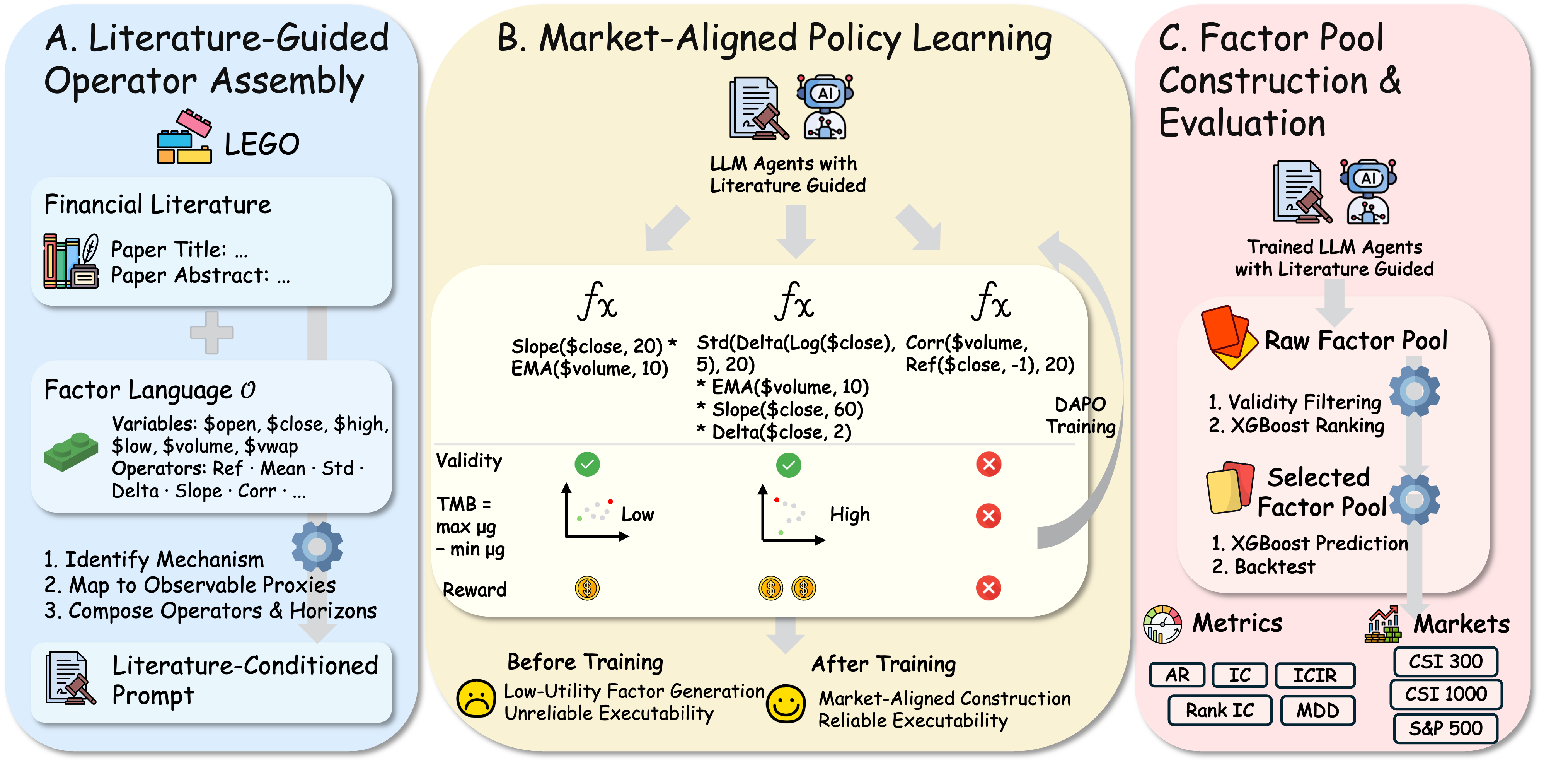}
    \caption{Overview of AlphaRLM. (A) Literature-Guided Operator Assembly (LEGO) combines paper titles and abstracts with permitted market variables and operators to guide factor construction. (B) Executable-validity and grouped-return separation rewards guide DAPO updates to the generation policy. (C) Generated factors undergo screening and training-set XGBoost feature-importance selection, followed by downstream return prediction and portfolio backtesting.}
    \label{fig:architecture}
\end{figure*}
Alpha factor discovery seeks predictive signals that map market observations to stock rankings \citep{yu_generating_2023}.
Economic theory has long guided factor design \citep{fama_french_1993,fama_five_factor_2015}, while automated discovery requires constructing expressions from market variables, operators, and temporal windows.
This creates a large combinatorial search space in which executable expressions may capture spurious or redundant relationships \citep{harvey_2016,feng_taming_2020}.
A central challenge is to learn how to construct useful factors, combining a meaningful direction for exploration with empirical evaluation of the resulting signals.

Early automated factor mining builds on genetic programming, which searches mathematical expressions through mutation and crossover \citep{allen_karjalainen_1999,potvin_2004}.
Subsequent methods extend evolutionary search and use reinforcement learning (RL) to learn expression-construction policies from market performance \citep{cui_alphaevolve_2021,yu_generating_2023}.
These approaches automate the exploration of candidate formulas, with historical evaluation guiding which expressions to retain or generate next.
These approaches organize search primarily around market variables and numerical objectives, while LLMs provide a complementary mechanism for incorporating textual economic knowledge into factor construction.

Large language models (LLMs) have broadened factor discovery by translating human investment ideas and financial research into executable candidate expressions \citep{wang2025alpha,kou2024automate,tang_alphaagent_2025}.
Recent systems improve this process through experience memory and trajectory evolution \citep{wang_factorminer_2026,han_quantaalpha_2026}, while RL-based approaches train LLMs to improve and diversify factors, primarily using seed expressions to structure learning \citep{tang_alphaagentevo_2026,zhang_quantevolver_2026}.
Despite this progress, direct factor construction through policy learning remains less explored, particularly when the model is not given an input factor expression to mutate or refine.
This setting requires the model to navigate a massive combinatorial space while discovering useful market signals.
The central challenge is therefore to provide informative domain priors for open-ended exploration together with a learnable market-based objective that can guide policy optimization under noisy and non-stationary feedback.

We propose AlphaRLM, a framework that addresses this challenge by coupling domain knowledge with market-aligned reinforcement learning (Figure~\ref{fig:architecture}).
Literature-Guided Operator Assembly (LEGO) conditions generation on economic ideas conveyed by academic papers and guides the model to map them onto available market variables and operators.
To learn from market feedback, we optimize the policy via Decoupled Clip and Dynamic Sampling Policy Optimization (DAPO) \citep{yu_dapo_2025}, employing a grouped-return separation reward.
This reward pools next-day return observations across dates within each factor-quantile group and measures the range of the resulting group means, allowing non-monotonic group-return patterns whose strongest contrast need not occur between the two factor-value extremes.
Together, these results demonstrate the potential of literature-conditioned policy learning for constructing executable factors with strong downstream utility.

Operating with an 8B model, AlphaRLM outperforms all evaluated baselines in annualized excess returns across CSI 300, CSI 1000, and S\&P 500. It notably surpasses both traditional search paradigms such as genetic programming and far larger LLM-based solutions.  Beyond portfolio performance, factor-level analyses show that AlphaRLM generates composite signals that exhibit strong grouped-return separation. Ultimately, by coupling domain literature with closed-loop market feedback, AlphaRLM synthesizes expressive composite signals that translate directly into superior downstream stock selection.

Our contributions are threefold.
\begin{itemize}
    \item We propose AlphaRLM, a reinforcement-learning framework for direct formulaic alpha discovery without relying on seed factors, incorporating Literature-Guided Operator Assembly (LEGO) to bridge unstructured financial literature with structured, executable symbolic factors.
    \item We design a market-aligned reward mechanism that combines executable validity with grouped-return separation (TMB), enabling the policy to learn from non-monotonic return structures under noisy market feedback.
    \item We evaluate AlphaRLM across three distinct equity markets (CSI 300, CSI 1000, and S\&P 500), demonstrating that the aligned 8B model consistently outperforms a range of strong baselines in downstream stock selection.
\end{itemize}

\section{Related Work}

\paragraph{Formulaic alpha discovery.}
Formulaic alpha discovery searches for interpretable expressions built from market variables and domain-specific operators.
Genetic programming explores this space through mutation and crossover \citep{allen_karjalainen_1999,potvin_2004}, while AlphaEvolve, AlphaGen, and AlphaForge use evolutionary or reinforcement-learning search guided by historical market performance \citep{cui_alphaevolve_2021,yu_generating_2023,shi_alphaforge_2024}.
QuantFactor REINFORCE further improves policy-gradient estimation and reward shaping for factor stability \citep{zhao_quantfactor_2024}.

\paragraph{LLM-driven alpha mining.}
LLMs extend alpha mining with textual knowledge and reasoning, as summarized in recent surveys \citep{zhang2025survey}.
LLMFactor extracts textual drivers of stock movements \citep{wang_llmfactor_2024}, while Alpha-GPT and Kou et al.\ translate investment ideas or financial research into executable candidate factors \citep{wang2025alpha,kou2024automate}.
Recent agentic systems combine generation with retrieval, experience memory, multi-agent reasoning, and iterative market evaluation \citep{tang_alphaagent_2025,li2026r,vu2026self,wang_factorminer_2026}.
AlphaJungle, QuantaAlpha, and CogAlpha refine formulas or mining trajectories through tree search and evolutionary mechanisms \citep{shi2026navigating,han_quantaalpha_2026,liu2026cognitive}, while AlphaSchema structures exploration around trading semantics \citep{yi2026alphaschema}.

\paragraph{Market feedback and policy learning.}
Recent work also incorporates reinforcement learning more directly into LLM-based financial reasoning and factor discovery.
Fin-R1 and Alpha-R1 apply RL to financial reasoning and factor screening \citep{liu_finr1_2026,jiang_alphar1_2025}, while HARLA places LLM generation inside an RL-based alpha optimization loop \citep{yu2026hybrid}.
AlphaAgentEvo and QuantEvolver further train LLM policies to evolve or complement existing factor expressions using market-derived feedback \citep{tang_alphaagentevo_2026,zhang_quantevolver_2026}.
AlphaRLM studies a complementary setting: \emph{direct factor construction through policy learning}, where financial literature provides semantic conditioning and the model learns to construct executable factors without an input factor expression to mutate or refine.

\section{Method}

AlphaRLM learns to construct executable alpha factors from economic ideas in financial literature.
Figure~\ref{fig:architecture} summarizes literature-guided operator assembly (A), market-aligned policy learning (B), and factor pool construction and evaluation (C).

\subsection{Problem Formulation}

We consider a universe of $n$ stocks and denote the market history available on trading day $t$ by $X_t$.
An alpha factor is a symbolic expression $f$ that maps this history to cross-sectional scores:
\begin{equation}
  v_t=f(X_t)\in\mathbb{R}^{n}.
\end{equation}
The expression language follows Qlib-style factors \citep{yang_qlib_2020}, combining market variables such as \texttt{\$open}, \texttt{\$close}, \texttt{\$high}, \texttt{\$low}, \texttt{\$volume}, and \texttt{\$vwap} with an operator library $\mathcal{O}$ containing operations such as \texttt{Ref}, \texttt{Mean}, \texttt{Std}, \texttt{Delta}, \texttt{Rank}, and \texttt{Corr}.

Let $s$ denote a literature inspiration consisting of a paper's title and abstract, and let $q(s,\mathcal{O})$ be its factor-generation prompt.
The policy $\pi_\theta$ produces a completion $o$, from which an extraction function $E$ obtains a candidate expression $f=E(o)$.
We learn the policy by maximizing expected reward over training inspirations:
\begin{equation}
 \max_\theta\;
 \mathbb{E}_{s\sim\mathcal{D}_{\mathrm{lit}}^{\mathrm{train}}}
 \mathbb{E}_{o\sim\pi_\theta(\cdot\mid q(s,\mathcal{O}))}
 \left[R(o;\mathcal{D}_{\mathrm{mkt}}^{\mathrm{train}})\right].
\end{equation}
Here, $\mathcal{D}_{\mathrm{mkt}}^{\mathrm{train}}$ contains historical market observations and subsequent returns for reward evaluation, and $R$ combines the factor task reward with DAPO's overlong-output penalty.
The learned policy maps literature-derived economic ideas to executable expressions over available market observables, which subsequently serve as features for downstream stock selection.

\subsection{Literature-Guided Operator Assembly}

LEGO conditions symbolic factor construction on economic ideas in finance papers and the expression language of the factor engine.
Variables and operators serve as modular building blocks assembled into executable expressions under guidance.
For each inspiration, the prompt combines the paper's title and abstract, permitted variables and operators, factor-mining instructions, and an output schema requesting reasoning, a factor expression, and a financial interpretation.
The literature provides semantic context, while the operator definitions specify the executable representation space.

The generation instructions guide the model to identify an economic mechanism, map it to observable proxies, and compose them through operators and temporal windows.
For example, a liquidity-related hypothesis must be expressed through available price and volume observables rather than an unavailable data series.
The grammar is supplied in the prompt, and generated expressions are subsequently checked by the factor engine.

Each completion provides a factor with reasoning and financial interpretation, while only the extracted expression is evaluated by the reward function.
Reasoning and output structure guide generation, whereas the reward measures expression validity and market usefulness; prose-format adherence is not separately rewarded.

\subsection{Market-Based Reward}

We evaluate each extracted factor using two rule-based rewards: executable validity and grouped-return separation.
Both are computed using the 2018--2019 market-data subset used for RL reward evaluation throughout training.

\paragraph{Executable validity.}
The validity reward $r_{\mathrm{valid}}(f)$ equals 1 when the factor can be parsed and computed without forward-looking feature access, and 0 otherwise.
Temporal causality is checked by extending the calculation window with future data and comparing factor values over the original window.
An expression whose historical values change under this extension is rejected.
This check evaluates the expression's use of market history, separately from the subsequent returns used to score its predictive usefulness.

\paragraph{Grouped-return separation.}
For each trading day $t$, stocks are independently sorted by factor value into $K$ quantile groups, with $K=100$ by default.
Let $B_{t,g}(f)$ denote the stocks assigned to group $g$, and let $y_{t+1,i}$ be stock $i$'s next-day return.
We pool stock--day return observations with the same group index across training dates $\mathcal{T}_{\mathrm{train}}$:
\begin{equation}
 \widehat{\mu}_g(f)=
 \frac{\displaystyle\sum_{t\in\mathcal{T}_{\mathrm{train}}}
       \sum_{i\in B_{t,g}(f)}y_{t+1,i}}
      {\displaystyle\sum_{t\in\mathcal{T}_{\mathrm{train}}}
       |B_{t,g}(f)|}.
\end{equation}
The top-minus-bottom spread (TMB) is computed from these pooled group means:
\begin{equation}
 \widehat{\Delta}_K(f)=\max_{g}\widehat{\mu}_g(f)-\min_{g}\widehat{\mu}_g(f).
\end{equation}
Thus, the spread compares the highest-return and lowest-return groups after temporal aggregation.
Unlike an endpoint spread that compares only the lowest and highest factor-value quantiles, TMB selects the most separated pair among all groups. It therefore captures endpoint separation while allowing the strongest return contrast to occur at other quantiles under non-monotonic factor--return relationships. Appendix~\ref{app:reward_analysis} provides a population interpretation, a comparison with endpoint separation, and an analysis of the group-resolution trade-off.

We scale the spread to obtain
\begin{equation}
r_{\mathrm{tmb}}(f)=
\begin{cases}
100\widehat{\Delta}_K(f), & \text{if valid},\\
0, & \text{otherwise}.
\end{cases}
\end{equation}
The task reward is
\begin{equation}
 r(f)=\alpha r_{\mathrm{valid}}(f)+\beta r_{\mathrm{tmb}}(f),
\end{equation}
where $\alpha$ and $\beta$ weight the two components.
The validity term provides a feasibility signal, while TMB differentiates valid factors by empirical grouped-return separation.

\subsection{Reinforcement Learning for Factor Generation}

We optimize the factor-generation policy with DAPO \citep{yu_dapo_2025}, which builds on the group-relative policy optimization introduced in DeepSeekMath \citep{shao_deepseekmath_2024}.
For each literature-conditioned prompt $q$, the rollout policy samples $G$ completions $\{o_i\}_{i=1}^{G}$, from which candidate factors $f_i$ are extracted.
Each candidate receives the task reward $r(f_i)$, together with DAPO's overlong-output penalty:
\begin{equation}
R_i = r(f_i) + r_{\mathrm{length}}(o_i).
\end{equation}

DAPO performs dynamic sampling, retaining prompt groups with nonzero reward variance, and computes group-relative advantages as
\begin{equation}
A_i =
\frac{
R_i-\mathrm{mean}(\{R_j\}_{j=1}^{G})
}{
\mathrm{std}(\{R_j\}_{j=1}^{G})+\epsilon
}.
\end{equation}
Because all completions in a group are conditioned on the same literature inspiration, this relative comparison provides a learning signal for how the same economic idea can be translated into different executable factor expressions.

For token position $k$ in completion $o_i$, DAPO defines the importance ratio and asymmetric clipping as
\begin{equation}
\begin{aligned}
\rho_{i,k}(\theta)
&=
\frac{
\pi_\theta(o_{i,k}\mid q,o_{i,<k})
}{
\pi_{\theta_{\mathrm{old}}}(o_{i,k}\mid q,o_{i,<k})
},\\
\bar{\rho}_{i,k}(\theta)
&=
\mathrm{clip}\!\left(
\rho_{i,k}(\theta),
1-\epsilon_{\mathrm{low}},
1+\epsilon_{\mathrm{high}}
\right).
\end{aligned}
\end{equation}

The resulting token-level policy objective is
\begin{equation}
\begin{aligned}
J_{\mathrm{DAPO}}(\theta)
&=
\mathbb{E}_{q,\{o_i\}\sim\pi_{\theta_{\mathrm{old}}}}
\Bigg[
\frac{1}{\sum_{i=1}^{G}L_i}
\sum_{i=1}^{G}
\sum_{k=1}^{L_i}
\\
&\qquad
\min\left(
\rho_{i,k}(\theta)A_i,
\bar{\rho}_{i,k}(\theta)A_i
\right)
\Bigg],
\end{aligned}
\label{eq:dapo_objective}
\end{equation}
where $L_i$ denotes the number of generated tokens in $o_i$.
Through this objective, the validity and grouped-return-separation rewards directly shape the generation policy toward executable factors with stronger empirical market signals.

Market feedback is used only as a training signal rather than conversational feedback during generation.
Unlike iterative backtest-guided factor evolution, the learned policy directly generates factors from literature-conditioned prompts without receiving intermediate evaluation results.
After training, the generated candidates are screened and selected for downstream stock prediction and portfolio evaluation, as described in Section~\ref{sec:experiments}.

\section{Experiments}
\label{sec:experiments}
\subsection{Experimental Setup}

\paragraph{Market data and splits.}
We collect daily stock data for CSI 300 and CSI 1000 constituents from Wind and S\&P 500 constituents from WRDS.
For downstream XGBoost modeling, all three markets use 2015--2019 for training, 2020 for validation, and 2021--2024 for testing.
To reduce the cost of evaluating generated factors during RL, training rewards are computed using the 2018--2019 subset of the market data for each policy's training universe.
Policy validation uses market data from 2020, while final backtests use the 2021--2024 test period.
CSI 300 and CSI 1000 cover large-cap and smaller-cap Chinese equities, respectively, while S\&P 500 covers U.S. equities.

\paragraph{Literature and models.}
For literature grounding, we retrieve papers from major finance, economics, management science, and related journals, supplemented with fintech-related arXiv papers.
We use keyword search and GPT-4-based relevance filtering to identify 5,524 finance-relevant papers from an initial pool of 8,765.
The main generation corpus contains 4,443 papers published during 2015--2019, split into 1,720 training prompts, 430 validation prompts, and 2,293 held-out generation prompts.
The publication-date restriction applies to all three literature splits.
We sample 2,000 non-finance papers published during 2015--2019 for domain-ablation experiments.
AlphaRLM uses Qwen3-8B and Qwen3-14B as base models.
For each model size, we train separate policies on CSI 300 and S\&P 500.
The CSI 300 policy generates a factor pool that is used unchanged for evaluation on both CSI 300 and CSI 1000, providing a test of transfer across stock universes.
The S\&P 500 policy generates the factors evaluated on S\&P 500.

\begin{table*}[t]
\centering
\small
\caption{
Backtest performance across CSI 300, CSI 1000, and S\&P 500.
Each cell reports mean $\pm$ standard deviation over five random seeds.
Within each dataset and metric, the best result is shown in \textbf{bold},
and the second-best result is \underline{underlined}.
}
\label{tab:main_results}

\renewcommand{\arraystretch}{1.18}
\setlength{\tabcolsep}{3pt}

\resizebox{\textwidth}{!}{
\begin{tabular}{lrrrrr|rrrrr|rrrrr}
\toprule
& \multicolumn{5}{c|}{\textbf{CSI 300}}
& \multicolumn{5}{c|}{\textbf{CSI 1000}}
& \multicolumn{5}{c}{\textbf{S\&P 500}} \\

Method
& AR & IC & RankIC & ICIR & MDD
& AR & IC & RankIC & ICIR & MDD
& AR & IC & RankIC & ICIR & MDD \\
\midrule

\rowcolor{gray!15}
\multicolumn{16}{c}{\textbf{\textit{Factor Libraries}}} \\

Alpha158
& \shortstack{0.0808\\{\scriptsize $\pm$ 0.0009}}
& \shortstack{0.0260\\{\scriptsize $\pm$ 0.0009}}
& \shortstack{0.0373\\{\scriptsize $\pm$ 0.0011}}
& \shortstack{\underline{0.1663}\\{\scriptsize $\pm$ 0.0041}}
& \shortstack{-0.1161\\{\scriptsize $\pm$ 0.0047}}
& \shortstack{0.0737\\{\scriptsize $\pm$ 0.0025}}
& \shortstack{0.0199\\{\scriptsize $\pm$ 0.0007}}
& \shortstack{0.0254\\{\scriptsize $\pm$ 0.0007}}
& \shortstack{0.1640\\{\scriptsize $\pm$ 0.0054}}
& \shortstack{-0.1424\\{\scriptsize $\pm$ 0.0116}}
& \shortstack{-0.0325\\{\scriptsize $\pm$ 0.0055}}
& \shortstack{-0.0027\\{\scriptsize $\pm$ 0.0024}}
& \shortstack{-0.0001\\{\scriptsize $\pm$ 0.0025}}
& \shortstack{-0.0227\\{\scriptsize $\pm$ 0.0208}}
& \shortstack{-0.4084\\{\scriptsize $\pm$ 0.0852}} \\

Alpha360
& \shortstack{0.0704\\{\scriptsize $\pm$ 0.0033}}
& \shortstack{0.0231\\{\scriptsize $\pm$ 0.0017}}
& \shortstack{0.0325\\{\scriptsize $\pm$ 0.0005}}
& \shortstack{0.1582\\{\scriptsize $\pm$ 0.0138}}
& \shortstack{-0.1022\\{\scriptsize $\pm$ 0.0099}}
& \shortstack{0.0703\\{\scriptsize $\pm$ 0.0007}}
& \shortstack{0.0206\\{\scriptsize $\pm$ 0.0013}}
& \shortstack{0.0263\\{\scriptsize $\pm$ 0.0007}}
& \shortstack{0.1676\\{\scriptsize $\pm$ 0.0113}}
& \shortstack{\underline{-0.1113}\\{\scriptsize $\pm$ 0.0050}}
& \shortstack{0.0160\\{\scriptsize $\pm$ 0.0060}}
& \shortstack{-0.0044\\{\scriptsize $\pm$ 0.0021}}
& \shortstack{-0.0058\\{\scriptsize $\pm$ 0.0009}}
& \shortstack{-0.0497\\{\scriptsize $\pm$ 0.0248}}
& \shortstack{-0.3162\\{\scriptsize $\pm$ 0.0688}} \\

\midrule
\rowcolor{gray!15}
\multicolumn{16}{c}{\textbf{\textit{Automated Factor Mining}}} \\

GP-Learn
& \shortstack{0.0815\\{\scriptsize $\pm$ 0.0013}}
& \shortstack{0.0185\\{\scriptsize $\pm$ 0.0016}}
& \shortstack{0.0271\\{\scriptsize $\pm$ 0.0018}}
& \shortstack{0.1181\\{\scriptsize $\pm$ 0.0093}}
& \shortstack{-0.1132\\{\scriptsize $\pm$ 0.0106}}
& \shortstack{0.0965\\{\scriptsize $\pm$ 0.0016}}
& \shortstack{0.0259\\{\scriptsize $\pm$ 0.0001}}
& \shortstack{0.0368\\{\scriptsize $\pm$ 0.0002}}
& \shortstack{0.1918\\{\scriptsize $\pm$ 0.0002}}
& \shortstack{-0.1727\\{\scriptsize $\pm$ 0.0194}}
& \shortstack{-0.1301\\{\scriptsize $\pm$ 0.0104}}
& \shortstack{-0.0016\\{\scriptsize $\pm$ 0.0015}}
& \shortstack{0.0028\\{\scriptsize $\pm$ 0.0011}}
& \shortstack{-0.0120\\{\scriptsize $\pm$ 0.0118}}
& \shortstack{-0.5591\\{\scriptsize $\pm$ 0.0838}} \\

AlphaGen
& \shortstack{0.0719\\{\scriptsize $\pm$ 0.0012}}
& \shortstack{0.0178\\{\scriptsize $\pm$ 0.0004}}
& \shortstack{0.0312\\{\scriptsize $\pm$ 0.0012}}
& \shortstack{0.1165\\{\scriptsize $\pm$ 0.0043}}
& \shortstack{-0.0888\\{\scriptsize $\pm$ 0.0110}}
& \shortstack{0.1088\\{\scriptsize $\pm$ 0.0008}}
& \shortstack{0.0290\\{\scriptsize $\pm$ 0.0006}}
& \shortstack{0.0367\\{\scriptsize $\pm$ 0.0002}}
& \shortstack{0.2316\\{\scriptsize $\pm$ 0.0045}}
& \shortstack{-0.1274\\{\scriptsize $\pm$ 0.0031}}
& \shortstack{-0.0327\\{\scriptsize $\pm$ 0.0023}}
& \shortstack{\textbf{0.0023}\\{\scriptsize $\pm$ 0.0010}}
& \shortstack{\underline{0.0051}\\{\scriptsize $\pm$ 0.0014}}
& \shortstack{\textbf{0.0178}\\{\scriptsize $\pm$ 0.0076}}
& \shortstack{-0.5236\\{\scriptsize $\pm$ 0.0272}} \\

AlphaForge
& \shortstack{0.0689\\{\scriptsize $\pm$ 0.0012}}
& \shortstack{0.0146\\{\scriptsize $\pm$ 0.0018}}
& \shortstack{0.0266\\{\scriptsize $\pm$ 0.0020}}
& \shortstack{0.0938\\{\scriptsize $\pm$ 0.0123}}
& \shortstack{-0.0922\\{\scriptsize $\pm$ 0.0141}}
& \shortstack{0.1190\\{\scriptsize $\pm$ 0.0017}}
& \shortstack{0.0222\\{\scriptsize $\pm$ 0.0008}}
& \shortstack{0.0322\\{\scriptsize $\pm$ 0.0004}}
& \shortstack{0.1648\\{\scriptsize $\pm$ 0.0063}}
& \shortstack{-0.1438\\{\scriptsize $\pm$ 0.0100}}
& \shortstack{-0.0790\\{\scriptsize $\pm$ 0.0091}}
& \shortstack{0.0007\\{\scriptsize $\pm$ 0.0015}}
& \shortstack{\textbf{0.0061}\\{\scriptsize $\pm$ 0.0007}}
& \shortstack{0.0053\\{\scriptsize $\pm$ 0.0112}}
& \shortstack{-0.4080\\{\scriptsize $\pm$ 0.0323}} \\

\midrule
\rowcolor{gray!15}
\multicolumn{16}{c}{\textbf{\textit{LLM-based Factor Mining}}} \\

DeepSeek-R1
& \shortstack{0.0856\\{\scriptsize $\pm$ 0.0062}}
& \shortstack{0.0237\\{\scriptsize $\pm$ 0.0019}}
& \shortstack{0.0380\\{\scriptsize $\pm$ 0.0002}}
& \shortstack{0.1632\\{\scriptsize $\pm$ 0.0131}}
& \shortstack{-0.0750\\{\scriptsize $\pm$ 0.0045}}
& \shortstack{0.1615\\{\scriptsize $\pm$ 0.0008}}
& \shortstack{0.0264\\{\scriptsize $\pm$ 0.0013}}
& \shortstack{0.0341\\{\scriptsize $\pm$ 0.0013}}
& \shortstack{0.2275\\{\scriptsize $\pm$ 0.0112}}
& \shortstack{\textbf{-0.1098}\\{\scriptsize $\pm$ 0.0180}}
& \shortstack{-0.1766\\{\scriptsize $\pm$ 0.0102}}
& \shortstack{-0.0016\\{\scriptsize $\pm$ 0.0007}}
& \shortstack{0.0035\\{\scriptsize $\pm$ 0.0008}}
& \shortstack{-0.0133\\{\scriptsize $\pm$ 0.0061}}
& \shortstack{-0.7385\\{\scriptsize $\pm$ 0.1087}} \\

GLM-5.2
& \shortstack{0.1007\\{\scriptsize $\pm$ 0.0006}}
& \shortstack{0.0219\\{\scriptsize $\pm$ 0.0015}}
& \shortstack{0.0335\\{\scriptsize $\pm$ 0.0021}}
& \shortstack{0.1398\\{\scriptsize $\pm$ 0.0089}}
& \shortstack{-0.0707\\{\scriptsize $\pm$ 0.0091}}
& \shortstack{0.1523\\{\scriptsize $\pm$ 0.0012}}
& \shortstack{0.0294\\{\scriptsize $\pm$ 0.0005}}
& \shortstack{0.0339\\{\scriptsize $\pm$ 0.0003}}
& \shortstack{\textbf{0.2417}\\{\scriptsize $\pm$ 0.0057}}
& \shortstack{-0.1429\\{\scriptsize $\pm$ 0.0115}}
& \shortstack{0.0219\\{\scriptsize $\pm$ 0.0059}}
& \shortstack{0.0018\\{\scriptsize $\pm$ 0.0005}}
& \shortstack{0.0017\\{\scriptsize $\pm$ 0.0017}}
& \shortstack{\underline{0.0170}\\{\scriptsize $\pm$ 0.0045}}
& \shortstack{-0.2145\\{\scriptsize $\pm$ 0.0310}} \\

Fin-R1
& \shortstack{0.0725\\{\scriptsize $\pm$ 0.0014}}
& \shortstack{\underline{0.0264}\\{\scriptsize $\pm$ 0.0005}}
& \shortstack{\underline{0.0411}\\{\scriptsize $\pm$ 0.0005}}
& \shortstack{\textbf{0.1695}\\{\scriptsize $\pm$ 0.0036}}
& \shortstack{-0.0683\\{\scriptsize $\pm$ 0.0052}}
& \shortstack{0.1072\\{\scriptsize $\pm$ 0.0030}}
& \shortstack{0.0289\\{\scriptsize $\pm$ 0.0004}}
& \shortstack{0.0377\\{\scriptsize $\pm$ 0.0009}}
& \shortstack{0.2317\\{\scriptsize $\pm$ 0.0051}}
& \shortstack{-0.1404\\{\scriptsize $\pm$ 0.0159}}
& \shortstack{-0.0624\\{\scriptsize $\pm$ 0.0065}}
& \shortstack{-0.0037\\{\scriptsize $\pm$ 0.0020}}
& \shortstack{-0.0022\\{\scriptsize $\pm$ 0.0021}}
& \shortstack{-0.0276\\{\scriptsize $\pm$ 0.0141}}
& \shortstack{-0.4904\\{\scriptsize $\pm$ 0.0324}} \\

FactorMiner
& \shortstack{0.0262\\{\scriptsize $\pm$ 0.0016}}
& \shortstack{0.0118\\{\scriptsize $\pm$ 0.0007}}
& \shortstack{0.0223\\{\scriptsize $\pm$ 0.0004}}
& \shortstack{0.0807\\{\scriptsize $\pm$ 0.0039}}
& \shortstack{-0.1811\\{\scriptsize $\pm$ 0.0126}}
& \shortstack{0.0971\\{\scriptsize $\pm$ 0.0018}}
& \shortstack{0.0213\\{\scriptsize $\pm$ 0.0002}}
& \shortstack{0.0380\\{\scriptsize $\pm$ 0.0003}}
& \shortstack{0.1640\\{\scriptsize $\pm$ 0.0006}}
& \shortstack{-0.1743\\{\scriptsize $\pm$ 0.0058}}
& \shortstack{0.0316\\{\scriptsize $\pm$ 0.0047}}
& \shortstack{-0.0021\\{\scriptsize $\pm$ 0.0008}}
& \shortstack{0.0018\\{\scriptsize $\pm$ 0.0006}}
& \shortstack{-0.0179\\{\scriptsize $\pm$ 0.0068}}
& \shortstack{-0.1732\\{\scriptsize $\pm$ 0.0816}} \\

QuantaAlpha
& \shortstack{0.0720\\{\scriptsize $\pm$ 0.0042}}
& \shortstack{0.0219\\{\scriptsize $\pm$ 0.0013}}
& \shortstack{0.0272\\{\scriptsize $\pm$ 0.0007}}
& \shortstack{0.1489\\{\scriptsize $\pm$ 0.0098}}
& \shortstack{-0.1090\\{\scriptsize $\pm$ 0.0059}}
& \shortstack{0.1427\\{\scriptsize $\pm$ 0.0035}}
& \shortstack{0.0274\\{\scriptsize $\pm$ 0.0019}}
& \shortstack{0.0369\\{\scriptsize $\pm$ 0.0012}}
& \shortstack{0.2150\\{\scriptsize $\pm$ 0.0161}}
& \shortstack{-0.1193\\{\scriptsize $\pm$ 0.0120}}
& \shortstack{-0.0857\\{\scriptsize $\pm$ 0.0050}}
& \shortstack{-0.0033\\{\scriptsize $\pm$ 0.0009}}
& \shortstack{0.0017\\{\scriptsize $\pm$ 0.0010}}
& \shortstack{-0.0281\\{\scriptsize $\pm$ 0.0075}}
& \shortstack{-0.3660\\{\scriptsize $\pm$ 0.0699}} \\

\textbf{AlphaRLM (8B)}
& \shortstack{\underline{0.1214}\\{\scriptsize $\pm$ 0.0011}}
& \shortstack{0.0256\\{\scriptsize $\pm$ 0.0015}}
& \shortstack{\textbf{0.0424}\\{\scriptsize $\pm$ 0.0007}}
& \shortstack{0.1586\\{\scriptsize $\pm$ 0.0091}}
& \shortstack{\underline{-0.0615}\\{\scriptsize $\pm$ 0.0173}}
& \shortstack{\underline{0.1745}\\{\scriptsize $\pm$ 0.0015}}
& \shortstack{\textbf{0.0302}\\{\scriptsize $\pm$ 0.0011}}
& \shortstack{\textbf{0.0414}\\{\scriptsize $\pm$ 0.0005}}
& \shortstack{\underline{0.2398}\\{\scriptsize $\pm$ 0.0104}}
& \shortstack{-0.1198\\{\scriptsize $\pm$ 0.0117}}
& \shortstack{\underline{0.0694}\\{\scriptsize $\pm$ 0.0065}}
& \shortstack{\underline{0.0021}\\{\scriptsize $\pm$ 0.0014}}
& \shortstack{0.0031\\{\scriptsize $\pm$ 0.0012}}
& \shortstack{0.0158\\{\scriptsize $\pm$ 0.0114}}
& \shortstack{\underline{-0.1484}\\{\scriptsize $\pm$ 0.0512}} \\

\textbf{AlphaRLM (14B)}
& \shortstack{\textbf{0.1338}\\{\scriptsize $\pm$ 0.0013}}
& \shortstack{\textbf{0.0264}\\{\scriptsize $\pm$ 0.0009}}
& \shortstack{0.0408\\{\scriptsize $\pm$ 0.0007}}
& \shortstack{0.1635\\{\scriptsize $\pm$ 0.0052}}
& \shortstack{\textbf{-0.0605}\\{\scriptsize $\pm$ 0.0099}}
& \shortstack{\textbf{0.1746}\\{\scriptsize $\pm$ 0.0018}}
& \shortstack{\underline{0.0297}\\{\scriptsize $\pm$ 0.0008}}
& \shortstack{\underline{0.0380}\\{\scriptsize $\pm$ 0.0007}}
& \shortstack{0.2307\\{\scriptsize $\pm$ 0.0083}}
& \shortstack{-0.1435\\{\scriptsize $\pm$ 0.0027}}
& \shortstack{\textbf{0.0808}\\{\scriptsize $\pm$ 0.0125}}
& \shortstack{0.0014\\{\scriptsize $\pm$ 0.0009}}
& \shortstack{0.0045\\{\scriptsize $\pm$ 0.0015}}
& \shortstack{0.0113\\{\scriptsize $\pm$ 0.0068}}
& \shortstack{\textbf{-0.1477}\\{\scriptsize $\pm$ 0.0564}} \\

\bottomrule
\end{tabular}
}
\end{table*}

\begin{figure*}[t]

    \centering

    \includegraphics[width=0.98\textwidth]{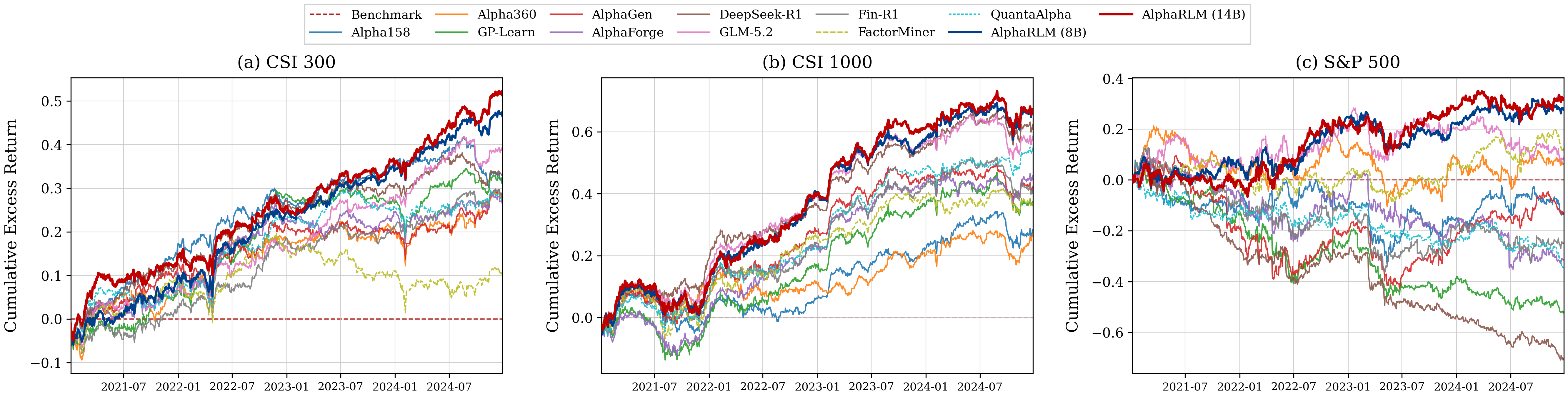}

    \caption{
    Cumulative excess return curves across CSI 300, CSI 1000, and S\&P 500.
    }

    \label{fig:backtest_curves}

\end{figure*}

\paragraph{Factor selection and baselines.}
Generated candidates undergo validity screening, followed by XGBoost feature-importance ranking on the training set to select 1,000 factors.
We compare three representative categories of methods: fixed factor libraries (Alpha158 and Alpha360 from Qlib \citep{yang_qlib_2020}); automated factor mining (GP-Learn, AlphaGen \citep{yu_generating_2023}, and AlphaForge \citep{shi_alphaforge_2024}); and LLM-based factor mining (DeepSeek-R1 \citep{guo_deepseekr1_2025}, GLM-5.2 \citep{glm_glm5_2026}, Fin-R1 \citep{liu_finr1_2026}, FactorMiner \citep{wang_factorminer_2026}, and QuantaAlpha \citep{han_quantaalpha_2026}).
The LLM baselines use the same generation budget and factor-selection procedure as AlphaRLM.
All methods use 1,000 factors for downstream evaluation except Alpha158 and Alpha360, which retain their factor libraries.

\paragraph{Backtesting and metrics.}
Generated factors are used as features in an XGBoost model \citep{chen_xgboost_2016} to predict next-day tradable returns.
This nonlinear predictor can capture interactions among factor features, motivated by evidence that nonlinear predictor interactions contribute to return predictability \citep{gu_empirical_2020}.
Features are robustly standardized with outlier clipping and missing-value imputation.
Returns are cross-sectionally rank-normalized.
The target is defined using future closing prices in a way that respects the T+1 trading rule.
Portfolios are rebalanced daily with a top-$k$ dropout strategy selecting the top 50 stocks by predicted return while excluding the bottom 5 from the candidate pool.
We exclude stocks that hit daily price limits in A-share markets and deduct transaction costs from portfolio returns.
All methods share the same XGBoost configuration, temporal splits, and portfolio strategy.
We report annualized excess return (AR), information coefficient (IC), rank information coefficient (RankIC), IC information ratio (ICIR), and maximum drawdown (MDD).
AR and MDD are computed from returns relative to the corresponding market index; IC and RankIC measure the correlation between predictions and realized returns, and ICIR measures the temporal stability of IC.
The main table aggregates results from five random seeds.
Each entry in Table~\ref{tab:main_results} reports the mean and standard deviation across these seeds.

\begin{table*}[t]
\centering
\small
\caption{Ablation results for AlphaRLM-8B on CSI 300 and S\&P 500.}
\label{tab:ablation}
\resizebox{0.8\textwidth}{!}{
\begin{tabular}{lrrrrrrrrrr}
\toprule
& \multicolumn{5}{c}{CSI 300} & \multicolumn{5}{c}{S\&P 500} \\
\cmidrule(lr){2-6}\cmidrule(lr){7-11}
Setting & AR\% & IC & RankIC & ICIR & MDD\% & AR\% & IC & RankIC & ICIR & MDD\% \\
\midrule
AlphaRLM (8B) & 12.14 & 0.026 & 0.042 & 0.159 & -6.15 & 6.94 & 0.002 & 0.003 & 0.016 & -14.84 \\
w/o DAPO & 9.66 & 0.025 & 0.034 & 0.141 & -7.58 & 3.14 & -0.001 & 0.004 & -0.004 & -30.72 \\
Non-financial text w/o DAPO & 5.24 & 0.013 & 0.035 & 0.115 & -12.99 & 1.78 & -0.003 & 0.004 & -0.022 & -26.46 \\
Non-financial text w/ DAPO & 2.13 & 0.009 & 0.031 & 0.068 & -17.59 & 2.90 & -0.008 & -0.002 & -0.063 & -28.72 \\
Linear reward & 9.68 & 0.031 & 0.042 & 0.184 & -15.60 & 0.77 & -0.002 & 0.004 & -0.028 & -19.88 \\
TMB 30 groups & 8.53 & 0.023 & 0.034 & 0.139 & -6.49 & 2.11 & 0.000 & 0.004 & -0.002 & -14.23 \\
\bottomrule
\end{tabular}
}
\end{table*}

\subsection{Main Results}

Table~\ref{tab:main_results} reports quantitative backtest performance across three equity universes, while Figure~\ref{fig:backtest_curves} shows the corresponding cumulative excess return trajectories.
Across all benchmarks, AlphaRLM achieves the two highest annualized excess returns (AR), with the 14B and 8B variants ranking first and second in every market.
Across all three markets, the factor pools discovered on CSI~300 transfer to the smaller-cap CSI~1000 universe without parameter updates, where both variants retain leading AR and competitive predictive correlations (IC and RankIC).
On the more efficient S\&P~500, where most baselines produce negative excess returns, AlphaRLM delivers positive AR of up to 8.08\% together with the smallest maximum drawdown ($-14.77\%$).
Notably, these return advantages are not accompanied by uniformly superior IC or RankIC, indicating that marginal predictive correlations alone do not fully explain downstream portfolio performance.
A similar pattern appears for AlphaGen and AlphaForge, which attain competitive correlation metrics on S\&P~500 but negative realized returns under the shared evaluation protocol.
Overall, the results show that AlphaRLM combines strong predictive signals with more favorable realized portfolio outcomes across distinct market-capitalization and geographic settings, while the 14B model provides modest AR gains over the 8B model in CSI~300 and S\&P~500.
The cumulative-return curves further show that these advantages persist over much of the 2021--2024 test period.

\paragraph{Summary.}
AlphaRLM consistently secures the top two AR positions across all three markets while providing superior drawdown control on CSI~300 and S\&P~500. Furthermore, the cross-universe results on CSI~1000 demonstrate strong out-of-distribution transferability, and the S\&P~500 results highlight its robust alignment with downstream execution over myopic statistical correlations. 

\subsection{Ablation Studies}

We analyze AlphaRLM across four perspectives: (1) core components, spanning literature grounding, DAPO, and reward design; (2) factor capacity under matched pool sizes; (3) sensitivity to the number of selected factors; and (4) sensitivity to the downstream predictor.

\paragraph{Component Ablations.}
Table~\ref{tab:ablation} isolates each core design choice on CSI~300 and S\&P~500.
Grounding generation in financial literature provides vital economic priors; replacing it with non-financial text severely hurts returns (AR on CSI~300: 12.14\%$\to$2.13\%) and exacerbates drawdowns.
With financial literature grounding, DAPO improves portfolio performance across both markets, most notably lifting AR (3.14\%$\to$6.94\%) and reducing MDD magnitude (30.72\%$\to$14.84\%) on S\&P~500.
The Linear reward ablation replaces the TMB component with an absolute endpoint spread,
\begin{equation}
r_{\mathrm{linear}}(f)=100\left|\widehat{\mu}_K(f)-\widehat{\mu}_1(f)\right|,
\end{equation}
where groups $1$ and $K$ denote the lowest and highest factor-value quantiles.
It uses $K=100$ and each market's 2018--2019 data, as does the default TMB reward.
TMB allows the strongest contrast to occur between any pair of groups.
Empirically, it yields a more favorable return--drawdown profile: Linear reward attains higher IC on CSI~300 but a larger drawdown magnitude (15.60\% versus 6.15\%), and lower S\&P~500 AR (0.77\% versus 6.94\%).
Reducing TMB to 30 groups lowers AR in both markets, although S\&P~500 MDD improves slightly ($-14.23\%$ versus $-14.84\%$), illustrating the empirical trade-off in group resolution.

\begin{figure}[h]
    \centering
    \includegraphics[width=\linewidth]{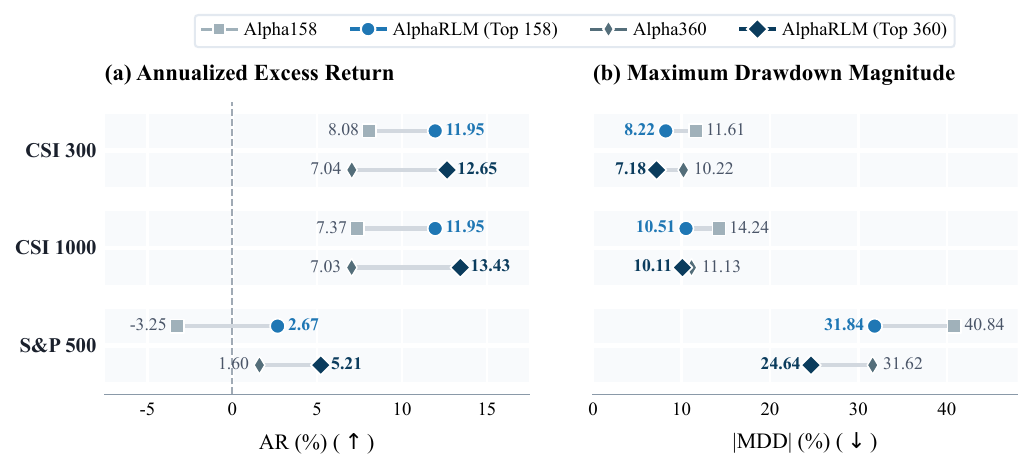}
    \caption{Matched factor-count comparison of AlphaRLM-8B with Alpha158 and Alpha360 across three markets. Each pair uses 158 or 360 factors.}
    \label{fig:matched_count}
\end{figure}

\paragraph{Matched Factor-Count Comparison.}
We compare subsets of 158 and 360 factors from AlphaRLM-8B with the corresponding fixed libraries, Alpha158 and Alpha360.
Figure~\ref{fig:matched_count} shows that, at both factor counts, AlphaRLM achieves higher annualized excess returns and smaller maximum drawdown magnitudes across all three markets.
These results show that its downstream performance advantage persists under matched factor counts.
Appendix~\ref{app:matched_count} provides the complete five-metric results and comparison details.

\paragraph{Choice of Factor Count.}
We further evaluate the setting of choosing 1,000 factors by varying the number of selected factors.
The results show a favorable return--drawdown balance at this setting, while larger pools provide no consistent improvement.
Appendix~\ref{app:factor_count} reports the full analysis.

\paragraph{Downstream Predictor Sensitivity.}
To disentangle factor-pool quality from nonlinear interactions learned by XGBoost, we replace only the downstream predictor with Ridge regression while keeping the selected factor pools and evaluation protocol fixed.
Despite this restricted linear aggregation, AlphaRLM retains the highest AR in every market through its 8B or 14B variant: 8.32\% on CSI~300, 9.58\% on CSI~1000, and $-3.90\%$ on S\&P~500.
Absolute performance is generally lower under Ridge, consistent with the limitations of a linear predictor in the low signal-to-noise setting of financial return prediction.
Together with the stronger factor-level TMB distributions in Section~\ref{sec:factor_tmb}, this provides complementary evidence that the generated factor pools themselves contain stronger predictive structure.
Appendix~\ref{app:ridge_sensitivity} provides the complete results.

\subsection{Factor-Level TMB Analysis}
\label{sec:factor_tmb}
\begin{figure}[h]
    \centering
    \includegraphics[width=0.46\textwidth]{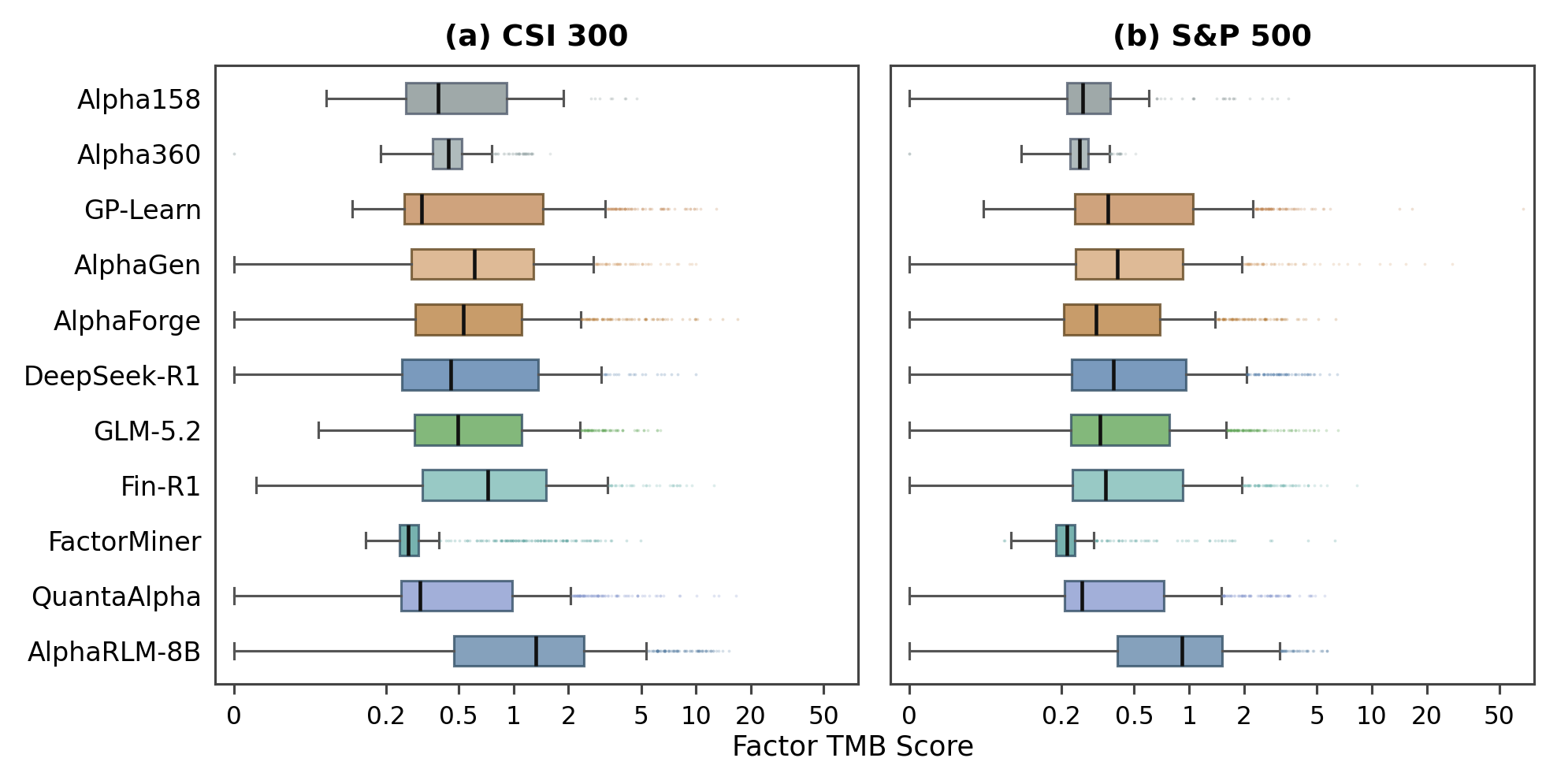}
    \caption{Factor-level TMB distributions on CSI 300 and S\&P 500. The horizontal axis is shown on a symmetric log scale to accommodate zero-valued factors.}
    \label{fig:tmb_distributions}
\end{figure}

We examine individual-factor TMB distributions on the 2021--2024 test period for CSI 300 and S\&P 500, using the final factor pools.
Figure~\ref{fig:tmb_distributions} compares these distributions across AlphaRLM-8B and the baseline factor pools.

AlphaRLM-8B achieves the highest median TMB among the compared methods in both markets.
Its first and third quartiles are also higher, indicating that the difference extends across the central factor distribution.
Several baselines exhibit large upper-tail values, but their median TMB remains lower.
The comparison therefore highlights AlphaRLM's stronger typical factor-level return separation across both markets.

These distributions provide factor-level evidence complementing AlphaRLM's downstream portfolio performance, while Appendix~\ref{sec:factor_redundancy} further examines behavioral and structural redundancy in the final pools.
The higher central TMB values are consistent with the grouped-return training objective.
The following section traces representative expressions back to their literature inputs.

\subsection{Literature-to-Factor Case Studies}
\label{sec:factor_cases}

To examine literature-conditioned factor construction, Table~\ref{tab:factor_cases} presents three factors from the frozen AlphaRLM (8B) A-share pool, each matched to its original paper-title-and-abstract input.

\begin{table}[h]
\centering
\small
\caption{Illustrative AlphaRLM (8B) factors with verified literature inputs.}
\label{tab:factor_cases}
\begingroup
\renewcommand{\arraystretch}{1.15}
\begin{tabular}{@{}p{\linewidth}@{}}
\toprule
\textbf{Factor 1: Quantile-Clipped Trading Activity per Price Movement} \\
\texttt{\scriptsize If(Gt(\$volume, Quantile(\$volume, 20, 0.95)),} \\
\texttt{\scriptsize \quad Quantile(\$volume, 20, 0.95),} \\
\texttt{\scriptsize \quad If(Lt(\$volume, Quantile(\$volume, 20, 0.05)),} \\
\texttt{\scriptsize \qquad Quantile(\$volume, 20, 0.05), \$volume))} \\
\texttt{\scriptsize \quad / Abs(Delta(\$close, 1))} \\
Literature input: \citet{Patzelt_2018}. \\
\midrule
\textbf{Factor 2: Downside Price-Change Variance with Dispersion Scaling} \\
\texttt{\scriptsize Div(Var(If(Gt(Sub(\$close, Ref(\$close, 1)), 0),} \\
\texttt{\scriptsize \quad 0, Sub(\$close, Ref(\$close, 1))), 20),} \\
\texttt{\scriptsize \quad Std(\$close, 20))} \\
Literature input: \citet{baruník2019totalasymmetricfrequencyconnectedness}. \\
\midrule
\textbf{Factor 3: Relative-Kurtosis Conditioned Trading Activity} \\
\texttt{\scriptsize Mul(Div(\$volume, Abs(Delta(\$close, 1))),} \\
\texttt{\scriptsize \quad Div(Kurt(\$close, 20),} \\
\texttt{\scriptsize \qquad Mean(Kurt(\$close, 20), 10)))} \\
Literature input: \citet{Graczyk_2016}. \\
\bottomrule
\end{tabular}
\endgroup
\end{table}

Factor~1 translates the volume--price impact relationship studied by \citet{Patzelt_2018} into clipped trading volume scaled by absolute price changes (IC = 0.0058, RankIC = 0.0175, TMB = 10.470).
Factor~2 adapts the asymmetric volatility perspective of \citet{baruník2019totalasymmetricfrequencyconnectedness} by isolating downside changes and measuring rolling variance relative to price dispersion (IC = $-0.0093$, RankIC = $-0.0276$, TMB = 6.210).
Factor~3 extends the higher-moment analysis of trading activity in \citet{Graczyk_2016} by combining trading volume with normalized price kurtosis (IC = 0.0030, RankIC = 0.0104, TMB = 2.413).
Together, these cases provide examples of how literature inputs induce different observable mappings and operator compositions.

\section{Conclusion}

We introduced AlphaRLM, a reinforcement-learning framework for direct formulaic alpha generation from financial literature without relying on seed factors. 
By combining Literature-Guided Operator Assembly (LEGO) with a market-aligned reward based on executable validity and grouped-return separation, AlphaRLM uses domain knowledge to guide open-ended exploration and empirical market feedback to align the generation of executable factors. 
In particular, grouped-return separation compares all factor quantiles rather than only the two endpoints, allowing the reward to capture non-monotonic factor--return relationships.
Across CSI 300, CSI 1000, and S\&P 500, the aligned models consistently outperform strong baselines in downstream stock selection, while ablations and factor-level analyses support the benefits of literature grounding, reinforcement learning, and the grouped-return reward design. 
These results highlight the potential of literature-grounded and market-aligned generation for transforming financial knowledge into structured, executable signals.

\section{Limitations}

While AlphaRLM demonstrates strong performance in symbolic factor mining, several directions remain for future work.
First, our evaluation focuses on daily equity data from Chinese and U.S. markets; extending the framework to intraday settings and other asset classes would provide broader validation.
Second, the literature-grounding corpus is intentionally broad, while future work could investigate more selective sources, such as top-tier finance and management journals, to study the effect of higher-quality domain priors.
Finally, due to the computational cost of reinforcement learning rollouts and symbolic verification, our experiments focus on Qwen3-8B and Qwen3-14B; evaluating larger and more diverse foundation models remains an important extension.

\section{Ethics Statement}

This work studies automated financial factor generation for research purposes and should not be interpreted as investment advice. 
The reported results are based on historical market data and therefore do not guarantee future investment performance. 
Any practical deployment should require independent validation, appropriate risk controls, compliance review, and human oversight. 
The literature prompts are constructed from paper titles and abstracts, and no private personal data are used. 

\bibliography{custom}

\clearpage
\appendix

\section{Data Construction}
\label{app:data}

\paragraph{Stock data.}
The equity experiments use daily bars from January 1, 2015 to December 31, 2024.
CSI 300 and CSI 1000 cover the Chinese A-share market, while S\&P 500 is used as a U.S. market generalization test.
All factors are constructed from historical open, high, low, close, volume, and VWAP-style price information.
Downstream XGBoost models are trained on 2015--2019 market data, whereas RL training rewards use the 2018--2019 subset to reduce evaluation cost.
Market data from 2020 are used for validation, and final backtests cover 2021--2024, after the 2019 literature cutoff.

\paragraph{Literature data.}
The literature repository is built from finance, economics, management science, and fintech-related sources.
The retrieval process has four stages: keyword construction, metadata retrieval, LLM-based relevance filtering, and publication-date filtering.
We first construct 45 finance-relevant search terms with broad coverage of asset pricing, factor investing, statistical arbitrage, machine learning in finance, and quantitative trading.
The initial retrieval returns 8,765 candidate papers.
An LLM relevance classifier is then applied to titles and abstracts, yielding 5,524 finance-relevant papers in the full corpus; 4,443 papers published during 2015--2019 are used for the main experiments.
These papers supply the training, validation, and held-out generation prompts under the same publication-date restriction.
For ablations, we sample 2,000 papers from the non-finance set under the same 2015--2019 publication-date restriction.

\paragraph{Retrieval keywords.}
The keywords are:
\emph{alpha factors, financial alpha generation, factor investing, quantitative alpha research, smart beta factors, momentum factor, mean reversion factor, multi-factor model, factor modeling in finance, risk factors in asset pricing, fundamental factors, quantitative finance alpha factors, factor decomposition, alpha factor analysis, alpha and beta in finance, machine learning alpha factors, alpha factor prediction, statistical arbitrage factors, portfolio construction with alpha factors, alpha generation and factor strategies, quantitative finance, algorithmic trading, financial modeling, risk management in quantitative finance, quantitative investment strategies, machine learning in finance, deep learning for financial prediction, statistical arbitrage, portfolio optimization, derivative pricing models, market microstructure, volatility modeling, high-frequency trading, asset pricing, factor investing in finance, financial econometrics, backtesting trading strategies, sentiment analysis in finance, time series analysis for finance, reinforcement learning in trading, quantitative risk assessment, quantitative asset management, quantitative trading models, financial anomaly detection, predictive analytics in finance}.

\section{Factor Language and Metrics}
\label{app:factor_language}

\paragraph{Symbolic factor language.}
Each generated factor must be an executable expression in a Qlib-style symbolic language.
The permitted variables include \texttt{\$open}, \texttt{\$close}, \texttt{\$high}, \texttt{\$low}, \texttt{\$volume}, and \texttt{\$vwap}.
The permitted operations include arithmetic composition, element-wise transforms such as \texttt{Abs}, \texttt{Sign}, and \texttt{Log}, and rolling time-series operators such as \texttt{Ref}, \texttt{Mean}, \texttt{Std}, \texttt{Max}, \texttt{Min}, \texttt{Delta}, \texttt{EMA}, \texttt{WMA}, \texttt{Slope}, and \texttt{Corr}.
For \texttt{Ref(x, d)}, valid generations must use nonnegative lags.
Negative lags would access future data and are rejected.

\paragraph{Annualized excess return.}
Let $r_t^{model}$ be the model portfolio return, $r_t^{bench}$ the benchmark return, $c_t$ the transaction cost, and $T$ the number of trading days.
We compute:
\begin{equation}
AR =
\left(\frac{1}{T}\sum_{t=1}^{T}
(r_t^{model}-r_t^{bench}-c_t)\right)\times 252.
\end{equation}

\paragraph{Information coefficient.}
The daily information coefficient is the Pearson correlation between predicted and realized returns:
\begin{equation}
IC_t=\mathrm{Corr}(\hat{\mathbf r}_t,\mathbf r_t).
\end{equation}
RankIC replaces Pearson correlation with Spearman rank correlation:
\begin{equation}
RankIC_t=\mathrm{Corr}_{\mathrm{Spearman}}(\hat{\mathbf r}_t,\mathbf r_t).
\end{equation}
ICIR measures temporal stability:
\begin{equation}
ICIR=\frac{\mathbb{E}_t[IC_t]}{\mathrm{Std}_t[IC_t]}.
\end{equation}

\paragraph{Maximum drawdown.}
For cumulative excess return
\begin{equation}
C_t=\sum_{i=1}^{t}(r_i^{model}-r_i^{bench}-c_i),
\end{equation}
drawdown and maximum drawdown are:
\begin{align}
DD(t)&=C_t-\max_{i\leq t}C_i,\\
MDD&=\min_{t\in[1,T]}DD(t).
\end{align}

\section{DAPO Training Details}
\label{app:grpo}

DAPO estimates the baseline from a group of rollouts instead of fitting a separate critic.
Given a prompt $q$, old policy $\pi_{\theta_{\mathrm{old}}}$, and $G$ generated outputs $\{o_i\}_{i=1}^{G}$, the clipped objective is:
\begin{equation}
\resizebox{.98\linewidth}{!}{$
\displaystyle
\hat{J}_{\mathrm{DAPO}}(\theta)
=\mathbb{E}\!\left[
\frac{1}{\sum_{j=1}^{G}|o_j|}\sum_{i=1}^{G}
\sum_{t=1}^{|o_i|}
\min\!\left(\rho_{i,t}A_i,
\mathrm{clip}(\rho_{i,t},1-\epsilon_{\mathrm{low}},1+\epsilon_{\mathrm{high}})A_i
\right)\right]$}
\end{equation}
where
\begin{equation}
\rho_{i,t}=
\frac{\pi_\theta(o_{i,t}\mid q,o_{i,<t})}
{\pi_{\theta_{\mathrm{old}}}(o_{i,t}\mid q,o_{i,<t})}.
\end{equation}
The relative advantage is normalized within the group:
\begin{equation}
A_i=\frac{R_i-\mathrm{mean}(\{R_j\}_{j=1}^{G})}
{\mathrm{std}(\{R_j\}_{j=1}^{G})+\epsilon}.
\end{equation}
Here $R_i$ includes the validity and TMB task rewards and the overlong-output penalty.
Dynamic sampling retains groups with nonzero total-reward variance and replenishes the batch before each update.

\begin{algorithm}[h]
\small
\caption{AlphaRLM DAPO Training}
\label{alg:app_grpo}
\begin{algorithmic}[1]
\Require Initial policy $\pi_{\theta}^{init}$, prompts $\mathcal{D}$, rewards $r_{\mathrm{valid}}$ and $r_{\mathrm{tmb}}$
\State $\pi_\theta \leftarrow \pi_{\theta}^{init}$
\For{each training step}
  \State Initialize an empty training buffer
  \While{the buffer is not full}
    \State Sample a prompt $q$ from $\mathcal{D}$
    \State Generate $G$ completions $\{o_i\}_{i=1}^{G}$
    \State Extract the final factor expression from each completion
    \State Compute task rewards and add overlong penalties
    \If{the group has nonzero total-reward variance}
      \State Normalize rewards and add the group to the buffer
    \EndIf
  \EndWhile
  \State Update $\pi_\theta$ with the clipped DAPO objective
\EndFor
\State \Return $\pi_\theta$
\end{algorithmic}
\end{algorithm}

\paragraph{Hardware and hyperparameters.}
Training uses 64 physical CPU cores and 8 NVIDIA A100-SXM4-80GB GPUs with bfloat16 mixed precision.
The configured training budget is 3 epochs, with a per-device batch size of 16, gradient accumulation steps of 1, a learning rate of $2\times10^{-6}$ with cosine scheduling, a warmup ratio of 0.01, a weight decay of 0.05, the AdamW optimizer. Following the decoupled clipping mechanism, the importance sampling clip ratio is decoupled into a lower bound of $\epsilon_{\text{low}} = 0.2$ and an upper bound of $\epsilon_{\text{high}} = 0.28$.
For both Qwen3-8B runs, the final checkpoint is used for factor generation.
Generation uses maximum prompt length 3700 tokens, maximum completion length 1600 tokens, temperature 1.0, and 8 completions per prompt.
The reward weights are $\alpha=1$ for validity and $\beta=4$ for TMB.

\paragraph{Validity reward.}
The validity reward checks whether a factor expression can be parsed and computed by the factor engine.
To detect future leakage, the calculation window is extended by 1 day, 1 week, 1 month, and 1 year.
If the original-window factor values change when future data are appended, the expression is considered forward-looking and receives zero validity reward.

\paragraph{TMB reward.}
For both the CSI~300 and S\&P~500 policies, the TMB training reward is computed on the corresponding market's 2018--2019 data to reduce the cost of rollout evaluation.
The system extracts the last matched factor expression from the completion, computes factor values, sorts stocks into 100 quantile groups, and measures the return spread between the best and worst groups.
The TMB spread is multiplied by 100 without clipping.
Invalid or unextractable factors receive zero TMB reward.

\paragraph{Output format.}
Reasoning and output structure are specified in the prompt; exact adherence to prose formatting is not a separate reward component.

\section{Reward Design and Analysis}
\label{app:reward_analysis}
\label{app:reward_rationale}

This section provides additional analysis of the market-based reward used
by AlphaRLM. Grouped-return separation supplies feedback for individual generated expressions without fitting a downstream predictor and running a portfolio backtest for every rollout. We analyze its predictive interpretation, its relation to the Linear reward ablation, and its interaction with DAPO.

Throughout this section, $K$ denotes the number of factor-value quantile
groups, while $G$ denotes the number of completions sampled from one
literature-conditioned prompt during DAPO training. Let $Y$ denote the next-day return used to compute the RL reward.

\subsection{Reward Decomposition and Predictive Interpretation}
\label{app:reward_predictive}

For a fixed factor $f$, let $B_{t,g}(f)$ denote the stocks assigned to
factor-value group $g\in\{1,\ldots,K\}$ on trading day $t$. The pooled
empirical mean return of group $g$ is
\begin{equation}
\widehat{\mu}_g(f)
=
\frac{
\sum_{t\in\mathcal{T}}
\sum_{i\in B_{t,g}(f)}
y_{t+1,i}
}{
\sum_{t\in\mathcal{T}}
|B_{t,g}(f)|
}.
\label{eq:empirical_group_mean_appendix}
\end{equation}
The empirical grouped-return separation is
\begin{equation}
\widehat{\Delta}_K(f)
=
\max_{1\le g\le K}\widehat{\mu}_g(f)
-
\min_{1\le g\le K}\widehat{\mu}_g(f).
\label{eq:empirical_tmb_appendix}
\end{equation}
The TMB component is $100\widehat{\Delta}_K(f)$ for valid factors, with $K=100$ by default.

For the population interpretation, consider a stock--day observation drawn from the reward-period population. Let $Z_f\in\{1,\ldots,K\}$ denote its group label under the factor's within-day cross-sectional quantile partition. The distribution mixes observations across dates, matching the observation-weighted empirical statistic. Assume $K\ge2$ and positive group probabilities. Define
\begin{equation}
p_g=\Pr(Z_f=g),
\qquad
\mu_g(f)=\mathbb{E}[Y\mid Z_f=g],
\end{equation}
and
\begin{equation}
\Delta_K(f)
=
\max_g\mu_g(f)-\min_g\mu_g(f).
\label{eq:population_tmb}
\end{equation}
We further define the between-group conditional-mean variation
\begin{equation}
\begin{aligned}
V_K(f)&=\operatorname{Var}\!\left(\mathbb{E}[Y\mid Z_f]\right)\\
&=\sum_{g=1}^{K}p_g\left(\mu_g(f)-\mathbb{E}[Y]\right)^2.
\end{aligned}
\label{eq:between_group_variation}
\end{equation}

\paragraph{Proposition 1 (Predictive meaning of the factor partition).}
Assume $\mathbb{E}[Y^2]<\infty$. Then
\begin{equation}
\begin{aligned}
V_K(f)&=\inf_{c\in\mathbb{R}}\mathbb{E}\!\left[(Y-c)^2\right]\\
&\quad-\inf_h\mathbb{E}\!\left[(Y-h(Z_f))^2\right].
\end{aligned}
\label{eq:mse_improvement}
\end{equation}
where the second infimum is over all functions
$h:\{1,\ldots,K\}\rightarrow\mathbb{R}$.

Moreover, if $p_{\min}=\min_g p_g>0$, then
\begin{equation}
\frac{p_{\min}}{2}\Delta_K(f)^2
\le
V_K(f)
\le
\frac{1}{4}\Delta_K(f)^2.
\label{eq:tmb_variance_bounds}
\end{equation}
For equal-probability quantile groups,
\begin{equation}
\frac{\Delta_K(f)^2}{2K}
\le
V_K(f)
\le
\frac{\Delta_K(f)^2}{4}.
\label{eq:tmb_variance_bounds_equal}
\end{equation}

\paragraph{Proof.}
The optimal constant predictor is $c^\star=\mathbb{E}[Y]$, whose
squared-error risk is $\operatorname{Var}(Y)$. The optimal predictor that
uses only the group label is
$h^\star(Z_f)=\mathbb{E}[Y\mid Z_f]$, whose risk is
$\mathbb{E}[\operatorname{Var}(Y\mid Z_f)]$.
Equation~\eqref{eq:mse_improvement} therefore follows from the law of total
variance.

If $\Delta_K(f)=0$, both bounds hold trivially. Otherwise, for the lower bound, let $a$ and $b$ index groups attaining the maximum
and minimum conditional means, and let $m=\mathbb{E}[Y]$. Then
\begin{align}
V_K(f)
&\ge
p_a(\mu_a-m)^2+p_b(\mu_b-m)^2 \\
&\ge
p_{\min}
\left[
(\mu_a-m)^2+(\mu_b-m)^2
\right] \\
&\ge
\frac{p_{\min}}{2}
(\mu_a-\mu_b)^2
=
\frac{p_{\min}}{2}\Delta_K(f)^2,
\end{align}
where the last inequality follows from
$x^2+y^2\ge (x-y)^2/2$, with $x=\mu_a-m$ and $y=\mu_b-m$.
For the upper bound, the random variable
$\mathbb{E}[Y\mid Z_f]$ lies in an interval of width $\Delta_K(f)$, and
the variance of any random variable supported on an interval of width
$\Delta$ is at most $\Delta^2/4$. This proves
Eq.~\eqref{eq:tmb_variance_bounds}.
\hfill$\square$

\paragraph{Interpretation.}
Equation~\eqref{eq:mse_improvement} shows that a factor-induced partition
is predictive when it reveals heterogeneous conditional mean returns.
TMB directly measures the range of these conditional means and therefore
serves as a simple surrogate for such predictive heterogeneity. The bounds
above do not imply that maximizing TMB is equivalent to maximizing
$V_K(f)$: TMB emphasizes the most separated pair of groups, whereas
$V_K(f)$ aggregates information across all groups.

\paragraph{Role of executable validity.}
Let $\chi(f)\in\{0,1\}$ indicate whether a generated expression is
extractable, executable, and passes the forward-looking check. Ignoring
DAPO's separate length penalty, the task reward can be written as
\begin{equation}
r(f)
=
\chi(f)
\left[
\alpha+100\beta\widehat{\Delta}_K(f)
\right].
\label{eq:task_reward_decomposition}
\end{equation}
For $\alpha>0$ and $\beta\ge0$, invalid expressions receive zero task
reward, whereas valid expressions receive at least $\alpha$.
The validity component thus acts as a feasibility signal, while the TMB
component differentiates feasible candidates according to empirical
grouped-return separation.

\subsection{Why Grouped-Return Separation?}
\label{app:reward_nonmonotonic}

The ablation labeled Linear reward in the main paper uses an absolute endpoint spread. Specifically, after sorting stocks into the same $K$ factor
quantiles and pooling returns across dates, it computes
\begin{equation}
\widehat{\Delta}_{\mathrm{end}}(f)
=
\left|
\widehat{\mu}_{K}(f)
-
\widehat{\mu}_{1}(f)
\right|,
\label{eq:endpoint_reward_empirical}
\end{equation}
where groups $1$ and $K$ are the lowest and highest factor-value
quantiles, respectively. The implemented reward is
$r_{\mathrm{linear}}(f)=100\widehat{\Delta}_{\mathrm{end}}(f)$. Both market-specific policies compute this reward on their respective 2018--2019 data.

In contrast, TMB searches over all factor-value groups:
\begin{equation}
\widehat{\Delta}_{K}(f)
=
\max_g\widehat{\mu}_g(f)
-
\min_g\widehat{\mu}_g(f).
\end{equation}

\paragraph{Proposition 2 (TMB generalizes endpoint separation).}
For any set of group means (population or empirical),
\begin{equation}
\Delta_{\mathrm{end}}(f)
=
|\mu_K(f)-\mu_1(f)|
\le
\Delta_K(f).
\label{eq:endpoint_dominated_by_tmb}
\end{equation}
Equality holds whenever the two endpoint groups attain the maximum and
minimum group means, possibly with ties.

\paragraph{Proof.}
Both $\mu_1(f)$ and $\mu_K(f)$ lie in the interval
$[\min_g\mu_g(f),\max_g\mu_g(f)]$. Therefore,
\begin{equation}
|\mu_K(f)-\mu_1(f)|
\le
\max_g\mu_g(f)-\min_g\mu_g(f).
\end{equation}
\hfill$\square$

Proposition~2 shows that TMB retains any separation captured by the
endpoint reward while relaxing the assumption that the strongest return
contrast must occur between the lowest and highest factor-value
quantiles. This matters when the factor--return relation is non-monotonic.

\paragraph{Proposition 3 (A non-monotonic signal missed by endpoint separation).}
Let
\begin{equation}
S\sim\operatorname{Unif}[-1,1],
\qquad
Y=S^2-\frac{1}{3}.
\end{equation}
For four equal-probability factor-value groups, the conditional mean
returns are
\begin{equation}
\left(
\frac{1}{4},
-\frac{1}{4},
-\frac{1}{4},
\frac{1}{4}
\right).
\label{eq:u_shape_group_means}
\end{equation}
Consequently,
\begin{equation}
\Delta_{\mathrm{end}}=0,
\qquad
\Delta_4=\frac{1}{2}.
\label{eq:u_shape_endpoint_vs_tmb}
\end{equation}
Moreover,
\begin{equation}
\operatorname{Corr}(S,Y)=0,
\qquad
\rho_{\mathrm{Spearman}}(S,Y)=0.
\label{eq:u_shape_correlations}
\end{equation}

\paragraph{Proof.}
The four factor-value intervals are
$[-1,-1/2]$, $[-1/2,0]$, $[0,1/2]$, and $[1/2,1]$.
Direct integration gives the conditional means in
Eq.~\eqref{eq:u_shape_group_means}, so the two endpoint groups have equal
mean return while the full grouped-return range is $1/2$.

By symmetry,
\begin{equation}
\operatorname{Cov}(S,Y)
=
\mathbb{E}\!\left[
S\left(S^2-\frac{1}{3}\right)
\right]
=
0,
\end{equation}
which gives zero Pearson correlation.
For continuous variables, Spearman correlation is the Pearson correlation
between their probability-integral transforms. Here,
\begin{equation}
F_S(S)=\frac{S+1}{2},
\qquad
F_Y(Y)=|S|.
\end{equation}
Symmetry gives $\mathbb{E}[S|S|]=0$, implying zero Spearman correlation.
\hfill$\square$

This example isolates the design motivation of TMB. An endpoint reward
can assign zero learning signal when the two factor-value extremes have
similar returns, even though intermediate and extreme groups differ
substantially. TMB instead asks whether \emph{any} factor-induced
quantile groups exhibit meaningful conditional-mean separation.

\paragraph{Scope.}
Grouped-return separation is not a universal dependence measure. It
captures heterogeneity in conditional mean returns. Signals expressed
only through conditional variance, or purely through interactions with
other features, may have small marginal TMB even when they are useful to
a multivariate downstream predictor. We therefore treat TMB as a
task-level surrogate rather than a complete characterization of factor
usefulness.

\subsection{Group Resolution and Temporal Aggregation}
\label{app:reward_resolution}

\paragraph{Resolution of the quantile partition.}
Let $T$ index the date and let $F_{f\mid T}$ be the within-day cross-sectional CDF of the factor score. Under continuous conditional score distributions,
\begin{equation}
U=F_{f\mid T}(f(X)\mid T)\in[0,1]
\end{equation}
is uniform conditional on each date and in the mixture across dates. This idealized population rank corresponds to the daily quantile sorting used in the reward. Define
\begin{equation}
m(u)=\mathbb{E}[Y\mid U=u].
\end{equation}
Let
\begin{equation}
\Delta_\infty
=
\sup_{u\in[0,1]}m(u)
-
\inf_{u\in[0,1]}m(u)
\end{equation}
be the continuous conditional-mean range, and let $\Delta_K$ denote the
range obtained after averaging $m(u)$ within $K$ equal-width rank
intervals.

\paragraph{Proposition 4 (Discretization error).}
If $m$ is $L$-Lipschitz on $[0,1]$, then
\begin{equation}
0
\le
\Delta_\infty-\Delta_K
\le
\frac{L}{K}.
\label{eq:resolution_bound}
\end{equation}

\paragraph{Proof.}
Each rank interval has width $1/K$. Let $u_{\max}$ maximize $m(u)$.
For the interval containing $u_{\max}$, the average distance from
$u_{\max}$ is at most $1/(2K)$. Lipschitz continuity therefore implies
that this interval's mean is at least
$\sup_u m(u)-L/(2K)$. Applying the same argument to a minimizer yields an
interval mean at most $\inf_u m(u)+L/(2K)$. Subtracting the two bounds
gives $\Delta_K\ge\Delta_\infty-L/K$. Since interval averages cannot
exceed the global maximum or fall below the global minimum,
$\Delta_K\le\Delta_\infty$.
\hfill$\square$

Finer grouping can therefore reduce discretization of localized
conditional-mean structure. It also reduces the number of observations
used to estimate each group mean.

\paragraph{Reference finite-sample bound.}
To make this trade-off explicit, consider an idealized fixed-factor
setting in which each group contains independent
$\sigma$-sub-Gaussian observations with mean $\mu_g$, and let
$n_{\min}$ be the smallest group sample size. A union bound over the $K$
group means yields, with probability at least $1-\delta$,
\begin{equation}
\left|
\widehat{\Delta}_K-\Delta_K
\right|
\le
2\sigma
\sqrt{
\frac{2\log(2K/\delta)}{n_{\min}}
}.
\label{eq:tmb_concentration}
\end{equation}
To see this, the sub-Gaussian mean bound and a union bound give
\begin{equation}
\begin{aligned}
&\Pr\!\left(\max_g|\widehat{\mu}_g-\mu_g|>u\right)\\
&\qquad\le 2K\exp\!\left(-\frac{n_{\min}u^2}{2\sigma^2}\right).
\end{aligned}
\end{equation}
The maximum and minimum each change by at most $\max_g|\widehat{\mu}_g-\mu_g|$, so
\begin{equation}
|\widehat{\Delta}_K-\Delta_K|
\le 2\max_g|\widehat{\mu}_g-\mu_g|.
\end{equation}
Setting $u=\sigma\sqrt{2\log(2K/\delta)/n_{\min}}$ yields Eq.~\eqref{eq:tmb_concentration}.

If $N$ observations are approximately balanced across groups,
$n_{\min}\approx N/K$, then combining
Eqs.~\eqref{eq:resolution_bound} and~\eqref{eq:tmb_concentration}
suggests the reference decomposition
\begin{equation}
\begin{aligned}
\left|\widehat{\Delta}_K-\Delta_\infty\right|
&\lesssim\underbrace{\frac{L}{K}}_{\text{discretization}}\\
&\quad+\underbrace{2\sigma\sqrt{\frac{2K\log(2K/\delta)}{N}}}_{\text{estimation}}.
\end{aligned}
\label{eq:resolution_estimation_tradeoff}
\end{equation}

Equation~\eqref{eq:resolution_estimation_tradeoff} is an interpretive
reference rather than a distribution-free guarantee for stock returns:
real market observations exhibit temporal and cross-sectional dependence,
and factors are generated adaptively during RL. It nevertheless makes the
group-resolution trade-off explicit. Coarser partitions reduce estimation
noise but can average away localized non-monotonic structure, whereas
finer partitions increase resolution at the cost of noisier group
estimates. Our 30-group versus 100-group ablation evaluates this trade-off for the current task.

\paragraph{Temporal pooling before the spread.}
The implemented reward first pools returns within the same group index
across dates and only then computes the maximum-minus-minimum spread. To
illustrate this ordering, let
$\mathbf{d}_t=(d_{t,1},\ldots,d_{t,K})$ collect group mean returns on date
$t$, and define
\begin{equation}
D(\mathbf{x})=\max_g x_g-\min_g x_g.
\end{equation}
If pooled group means use common nonnegative date weights
$\{w_t\}$ with $\sum_t w_t=1$, then
\begin{equation}
D\!\left(
\sum_t w_t\mathbf{d}_t
\right)
\le
\sum_t w_t D(\mathbf{d}_t).
\label{eq:pooling_range}
\end{equation}
Indeed,
$\max_g\sum_t w_t d_{t,g}\le\sum_t w_t\max_g d_{t,g}$ and
$\min_g\sum_t w_t d_{t,g}\ge\sum_t w_t\min_g d_{t,g}$.

Thus, group-level return patterns that reverse across dates can partially
cancel before the reward is computed, rather than being rewarded by
selecting a different best and worst group independently on every date.
In the implemented statistic, pooling is observation-weighted; the common
date-weight condition is exact when effective group counts induce the same
date weights across groups, and otherwise
Eq.~\eqref{eq:pooling_range} should be viewed as an explanatory special
case rather than a general bound for unequal group-specific date weights.

\subsection{Interaction with DAPO}
\label{app:reward_dapo}

AlphaRLM samples multiple completions from the same
literature-conditioned prompt and applies DAPO's group-relative
normalization. This gives a task-specific interpretation to the reward:
the model compares different symbolic realizations of the same
literature-derived economic idea.

Let
\begin{equation}
d_i=\widehat{\Delta}_K(f_i)
\end{equation}
for completion $i$. Consider a prompt group in which all extracted factors
are valid, $\beta>0$, and DAPO's overlong-output penalty is the same
constant $c$ for all completions. Then
\begin{equation}
R_i
=
\alpha+100\beta d_i+c.
\end{equation}

\paragraph{Proposition 5 (Relative optimization within a literature prompt).}
Under the conditions above, DAPO's normalized group-relative advantage is
\begin{equation}
A_i
=
\frac{
d_i-\overline{d}
}{
s_d+\epsilon/(100\beta)
},
\label{eq:tmb_relative_advantage}
\end{equation}
where $\overline{d}$ and $s_d$ are the mean and standard deviation of
$\{d_j\}_{j=1}^{G}$ within the prompt group.

\paragraph{Proof.}
Substituting
$R_i=\alpha+100\beta d_i+c$
into the group-relative advantage gives
\begin{align}
A_i
&=
\frac{
100\beta(d_i-\overline{d})
}{
100\beta s_d+\epsilon
} \\
&=
\frac{
d_i-\overline{d}
}{
s_d+\epsilon/(100\beta)
}.
\end{align}
The constants $\alpha$ and $c$, shared within the prompt group, cancel under centering.
\hfill$\square$

Equation~\eqref{eq:tmb_relative_advantage} shows that, within an all-valid
equal-penalty prompt group, the policy update compares alternative
symbolic constructions by their relative grouped-return separation. This
is particularly natural for AlphaRLM because all completions in the group
share the same literature inspiration: literature provides a common
semantic anchor, while market feedback differentiates how effectively
that idea is instantiated as an executable factor.

When a prompt group contains invalid outputs or different length
penalties, the validity and length terms also affect relative advantages.
Thus, Eq.~\eqref{eq:tmb_relative_advantage} is an interpretation of the
all-valid equal-penalty case.

\paragraph{Summary.}
The reward design separates two roles. Executable validity rewards feasible symbolic expressions, while grouped-return
separation supplies an empirical signal for conditional-mean differences
that need not occur between the two endpoint factor quantiles or follow a
globally monotonic pattern. Group resolution controls a
resolution--estimation trade-off, and DAPO uses the resulting market
feedback to compare alternative factor constructions conditioned on the
same literature-derived idea. These properties motivate TMB as a
task-level surrogate for factor generation.

\section{DAPO Training Dynamics}
\label{app:dapo_dynamics}

\begin{figure*}[t]
    \centering
    \includegraphics[width=\textwidth]{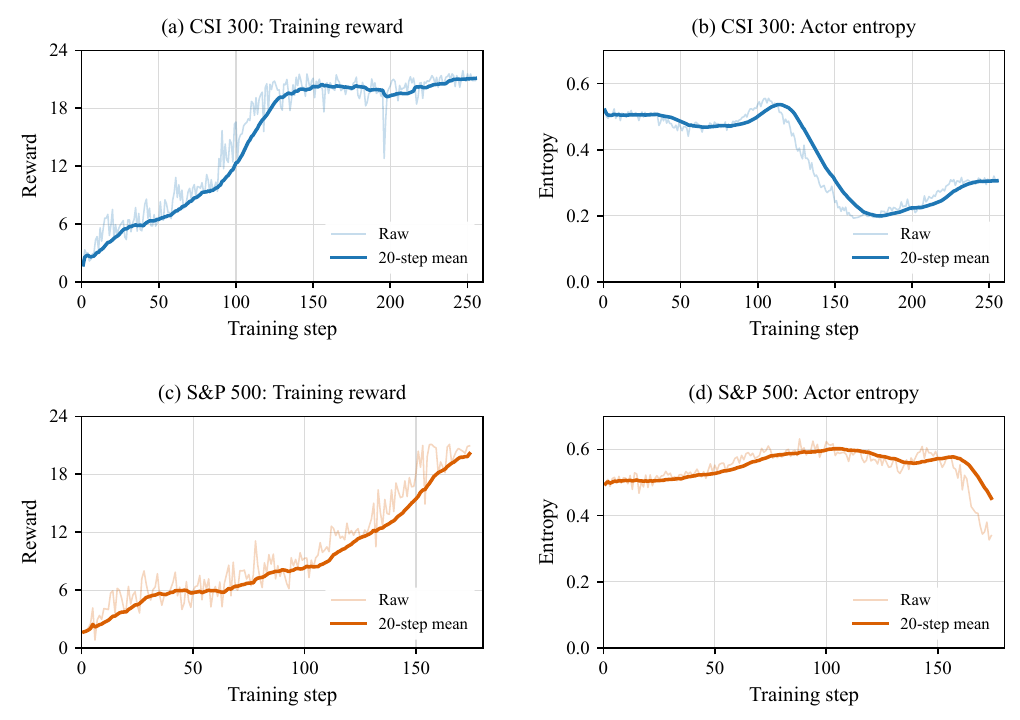}
    \caption{DAPO training dynamics for Qwen3-8B on CSI~300 and S\&P~500.}
    \label{fig:app_grpo_curves}
\end{figure*}

Figure~\ref{fig:app_grpo_curves} tracks training reward and actor entropy during the first training epoch for the Qwen3-8B policies trained separately on CSI~300 and S\&P~500.
The reward combines executable validity, weighted TMB, and DAPO's overlong-output penalty, with market-based feedback computed over 2018--2019.
Despite fluctuations in individual updates, the smoothed reward increases in both markets.
On CSI~300, it rises rapidly during the first half of training and becomes relatively stable in the later updates.
On S\&P~500, it increases more gradually at first and rises more sharply toward the end of the recorded run.
In both markets, the smoothed reward eventually converges to approximately 21.

Actor entropy follows distinct trajectories across the two markets.
On CSI~300, it decreases markedly during the middle of training and partially recovers later while the reward remains high.
On S\&P~500, entropy initially increases and falls near the end as the reward rises.

\section{Backtesting and Factor Filtering}
\label{app:backtest}

\begin{table}[h]
\centering
\small
\caption{Market-data splits for downstream XGBoost modeling. S\&P 500 trading-day counts follow the U.S. equity trading calendar.}
\label{tab:app_data_split}
\resizebox{\linewidth}{!}{
\begin{tabular}{lll}
\toprule
Asset & Period & Trading days \\
\midrule
\multirow{3}{*}{CSI 300}
& Training: 2015-01 to 2019-12 & 1219 \\
& Validation: 2020-01 to 2020-12 & 243 \\
& Testing: 2021-01 to 2024-12 & 968 \\
\midrule
\multirow{3}{*}{CSI 1000}
& Training: 2015-01 to 2019-12 & 1219 \\
& Validation: 2020-01 to 2020-12 & 243 \\
& Testing: 2021-01 to 2024-12 & 968 \\
\midrule
\multirow{3}{*}{S\&P 500}
& Training: 2015-01 to 2019-12 & 1258 \\
& Validation: 2020-01 to 2020-12 & 253 \\
& Testing: 2021-01 to 2024-12 & 1005 \\
\bottomrule
\end{tabular}
}
\end{table}

\paragraph{Downstream model.}
All methods are evaluated with the same XGBoost configuration: \texttt{eval\_metric=rmse}, \texttt{colsample\_bytree=0.8}, \texttt{eta=0.04}, \texttt{max\_depth=8}, \texttt{n\_estimators=600}, and \texttt{subsample=0.8}.
Features are robustly standardized with outlier clipping and missing-value imputation.
Targets are cross-sectionally rank-normalized next-day tradable returns, and samples with missing labels are removed.

\paragraph{Order execution.}
The backtest uses a daily top-$k$ dropout strategy.
The model selects the top 50 stocks by predicted return and excludes the bottom 5 from the candidate pool.
Trades are executed at closing prices, and only tradable stocks are included.
For A-share markets, stocks hitting daily limit-up or limit-down thresholds ($\pm$9.5\%) are excluded.
Transaction costs are 0.05\% for opening positions and 0.15\% for closing positions, with a minimum fee of 5 CNY per transaction.
The backtest runs from January 1, 2021 to December 30, 2024 with initial capital of 100 million CNY.

\paragraph{Factor filtering.}
The generated factor pool is filtered in two stages.
First, factor-construction validation removes syntactically invalid expressions and expressions with future leakage.
Second, XGBoost feature importance computed on the training set ranks the remaining factors, and the top 1,000 are retained.
The top158 and top360 variants use the same ranking and candidate pool, changing only the number of selected features.
The CSI 300-trained policy supplies the same factor pool for CSI 300 and CSI 1000 evaluation, while the S\&P 500-trained policy supplies the U.S. factor pool.

\begin{table*}[t]
\centering
\small
\caption{Complete results at matched factor counts. Generated subsets are derived from AlphaRLM-8B. AR and MDD are reported in percentages; fixed-library baselines are reproduced from Table~\ref{tab:main_results}.}
\label{tab:matched_count}
\begin{tabular}{llrrrrrr}
\toprule
Market & Factor pool & Count & AR & IC & RankIC & ICIR & MDD \\
\midrule
CSI 300 & Alpha158 & 158 & 8.08 & 0.0260 & 0.0373 & 0.1663 & -11.61 \\
 & AlphaRLM-top158 & 158 & 11.95 & 0.0239 & 0.0433 & 0.1478 & -8.22 \\
 & Alpha360 & 360 & 7.04 & 0.0231 & 0.0325 & 0.1582 & -10.22 \\
 & AlphaRLM-top360 & 360 & 12.65 & 0.0296 & 0.0470 & 0.1840 & -7.18 \\
\midrule
CSI 1000 & Alpha158 & 158 & 7.37 & 0.0199 & 0.0254 & 0.1640 & -14.24 \\
 & AlphaRLM-top158 & 158 & 11.95 & 0.0327 & 0.0384 & 0.2568 & -10.51 \\
 & Alpha360 & 360 & 7.03 & 0.0206 & 0.0263 & 0.1676 & -11.13 \\
 & AlphaRLM-top360 & 360 & 13.43 & 0.0327 & 0.0376 & 0.2606 & -10.11 \\
\midrule
S\&P 500 & Alpha158 & 158 & -3.25 & -0.0027 & -0.0001 & -0.0227 & -40.84 \\
 & AlphaRLM-top158 & 158 & 2.67 & 0.0023 & 0.0052 & 0.0182 & -31.84 \\
 & Alpha360 & 360 & 1.60 & -0.0044 & -0.0058 & -0.0497 & -31.62 \\
 & AlphaRLM-top360 & 360 & 5.21 & 0.0015 & 0.0047 & 0.0114 & -24.64 \\
\bottomrule
\end{tabular}
\end{table*}

\section{Matched Factor-Count Comparison}
\label{app:matched_count}

We compare AlphaRLM-8B subsets containing 158 and 360 factors with Alpha158 and Alpha360, respectively, on CSI~300, CSI~1000, and S\&P~500.
The top158 and top360 subsets use the same candidate pool and training-period XGBoost importance ranking as the default 1,000-factor configuration, changing only the number of retained factors.
The downstream model and backtesting protocol follow Appendix~\ref{app:backtest}, and results use the same five-seed aggregation as the main experiment.
Table~\ref{tab:matched_count} reports all five metrics and supplies the values plotted in Figure~\ref{fig:matched_count}.
MDD retains its signed convention in the table; the figure displays its absolute magnitude.

Across all six market--count combinations, AlphaRLM achieves higher AR and smaller drawdown magnitudes than the corresponding fixed library.
The correlation metrics do not improve uniformly: on CSI~300 at 158 factors, AlphaRLM has lower IC and ICIR than Alpha158, while its RankIC is higher.
Thus, the matched-count comparison establishes a consistent advantage in downstream return and drawdown across the evaluated settings.
Appendix~\ref{app:factor_count} complements this comparison by varying the number of factors selected from the AlphaRLM pool.

\section{Factor Redundancy and Structural Diversity}
\label{sec:factor_redundancy}

\subsection{Pairwise Factor Redundancy}

\paragraph{Evaluation protocol.}
We first examine whether strong downstream performance is accompanied by
excessive redundancy within the final factor pools.
For each market, we evaluate the final factor pools on the 2021--2024 test
period as a post-hoc diagnostic; these statistics are not used for factor
selection, model selection, or hyperparameter tuning.
For each pair of factors $(f_i,f_j)$, we pool all stock--day observations
for which both factors are valid over the full test period and compute
their Pearson correlation:
\begin{equation}
c_{ij}
=
\left|
\operatorname{Corr}
\left(
f_i(X),f_j(X)
\right)
\right|,
\qquad i<j.
\end{equation}
We take the absolute value because both strongly positive and strongly
negative correlations indicate linear redundancy.
Table~\ref{tab:factor_redundancy} reports the mean, standard deviation,
median, 95th percentile (P95), and the proportion of highly correlated
pairs satisfying $|r|\geq 0.8$.

\begin{table*}[t]
\centering
\small
\caption{
Pairwise absolute Pearson correlations within the final factor pools on the 2021--2024 test period.
For each factor pair, correlation is computed by pooling all jointly valid stock--day observations over the full test period.
$P_{0.8}$ denotes the proportion of factor pairs satisfying $|r|\geq 0.8$.
Lower values generally indicate less linear redundancy.
}
\label{tab:factor_redundancy}

\renewcommand{\arraystretch}{1.18}
\setlength{\tabcolsep}{3pt}

\resizebox{\textwidth}{!}{
\begin{tabular}{lrrrrr|rrrrr|rrrrr}
\toprule
& \multicolumn{5}{c|}{\textbf{CSI 300}}
& \multicolumn{5}{c|}{\textbf{CSI 1000}}
& \multicolumn{5}{c}{\textbf{S\&P 500}} \\

Method
& Mean & SD & Median & P95 & $P_{0.8}$
& Mean & SD & Median & P95 & $P_{0.8}$
& Mean & SD & Median & P95 & $P_{0.8}$ \\
\midrule

\rowcolor{gray!15}
\multicolumn{16}{c}{\textbf{\textit{Factor Libraries}}} \\

Alpha158
& 0.2638 & 0.2316 & 0.1924 & 0.7225 & 2.36\%
& 0.2651 & 0.2299 & 0.1976 & 0.7215 & 2.45\%
& 0.1680 & 0.2224 & 0.0742 & 0.6902 & 2.85\% \\

Alpha360
& 0.4887 & 0.3176 & 0.5111 & 0.9555 & 22.75\%
& 0.4985 & 0.3139 & 0.5337 & 0.9549 & 23.03\%
& 0.4747 & 0.3568 & 0.5295 & 0.9681 & 25.52\% \\

\midrule
\rowcolor{gray!15}
\multicolumn{16}{c}{\textbf{\textit{Automated Factor Mining}}} \\

GP-Learn
& 0.2819 & 0.2461 & 0.1898 & 0.8223 & 5.29\%
& 0.2641 & 0.2333 & 0.1832 & 0.7388 & 3.47\%
& 0.1793 & 0.2157 & 0.0830 & 0.7062 & 2.89\% \\

AlphaGen
& 0.1762 & 0.2184 & 0.0842 & 0.6777 & 2.39\%
& 0.1830 & 0.2170 & 0.0945 & 0.6895 & 2.31\%
& 0.1265 & 0.1892 & 0.0439 & 0.5745 & 1.50\% \\

AlphaForge
& 0.1769 & 0.2459 & 0.0564 & 0.7599 & 3.93\%
& 0.1543 & 0.2319 & 0.0372 & 0.7185 & 3.22\%
& 0.1024 & 0.1789 & 0.0245 & 0.5258 & 1.52\% \\

\midrule
\rowcolor{gray!15}
\multicolumn{16}{c}{\textbf{\textit{LLM-based Factor Mining}}} \\

DeepSeek-R1
& 0.0919 & 0.1265 & 0.0435 & 0.3585 & 0.17\%
& 0.1031 & 0.1279 & 0.0567 & 0.3715 & 0.15\%
& 0.0586 & 0.1093 & 0.0153 & 0.2814 & 0.12\% \\

GLM-5.2
& 0.1199 & 0.1358 & 0.0703 & 0.4081 & 0.11\%
& 0.1230 & 0.1348 & 0.0750 & 0.4074 & 0.08\%
& 0.0704 & 0.1091 & 0.0256 & 0.3016 & 0.07\% \\

Fin-R1
& 0.1185 & 0.1661 & 0.0502 & 0.4946 & 0.86\%
& 0.1017 & 0.1372 & 0.0472 & 0.3869 & 0.32\%
& 0.0679 & 0.1316 & 0.0138 & 0.3545 & 0.39\% \\

FactorMiner
& 0.2254 & 0.1931 & 0.1634 & 0.6272 & 0.97\%
& 0.2746 & 0.2004 & 0.2293 & 0.6713 & 1.30\%
& 0.1754 & 0.2257 & 0.0589 & 0.6880 & 2.00\% \\

QuantaAlpha
& 0.0708 & 0.1344 & 0.0193 & 0.3340 & 0.68\%
& 0.0679 & 0.1278 & 0.0184 & 0.2950 & 0.65\%
& 0.0475 & 0.1178 & 0.0064 & 0.2669 & 0.46\% \\

\textbf{AlphaRLM (8B)}
& 0.1093 & 0.1540 & 0.0467 & 0.4496 & 0.47\%
& 0.1214 & 0.1568 & 0.0597 & 0.4616 & 0.44\%
& 0.0558 & 0.1352 & 0.0049 & 0.3543 & 0.55\% \\

\textbf{AlphaRLM (14B)}
& 0.1044 & 0.1548 & 0.0396 & 0.4468 & 0.55\%
& 0.1163 & 0.1563 & 0.0532 & 0.4586 & 0.46\%
& 0.0595 & 0.1109 & 0.0144 & 0.2875 & 0.15\% \\

\bottomrule
\end{tabular}
}
\end{table*}

\begin{table*}[t]
\centering
\small
\caption{
Within-pool structural diversity and overlap with the standard Alpha158/Alpha360 libraries.
Exact uniqueness is the proportion of distinct exact canonical expressions.
Structural uniqueness is the proportion of distinct parameter-normalized skeletons.
Repeated-factor ratio denotes the fraction of factors belonging to skeleton clusters of size greater than one.
Library structural overlap is computed over unique AlphaRLM skeletons.
}
\label{tab:factor_structural_diversity}

\renewcommand{\arraystretch}{1.15}
\setlength{\tabcolsep}{5pt}

\resizebox{\textwidth}{!}{
\begin{tabular}{lrrrrrr}
\toprule
Pool
& Exact uniq.
& Structural uniq.
& Repeated factors
& Max cluster
& Exact library overlap
& Structural library overlap \\
\midrule

AlphaRLM-8B A-share
& 99.9\%
& 95.9\%
& 7.2\%
& 5
& 0.000\%
& 0.104\% \\

AlphaRLM-8B S\&P 500
& 99.5\%
& 89.1\%
& 18.4\%
& 10
& 0.500\%
& 0.337\% \\

AlphaRLM-14B A-share
& 99.7\%
& 80.7\%
& 28.1\%
& 22
& 0.100\%
& 0.124\% \\

AlphaRLM-14B S\&P 500
& 99.9\%
& 92.9\%
& 11.9\%
& 9
& 0.000\%
& 0.000\% \\

\bottomrule
\end{tabular}
}
\end{table*}

\paragraph{Results.}
AlphaRLM maintains low pairwise redundancy across all three markets while
retaining strong downstream performance.
For AlphaRLM-8B, the mean absolute correlation is 0.109 on CSI 300,
0.121 on CSI 1000, and 0.056 on S\&P 500; AlphaRLM-14B obtains
0.104, 0.116, and 0.060, respectively.
The corresponding medians range from only 0.005 to 0.060, indicating that
a typical pair of AlphaRLM factors exhibits weak linear dependence.
Both AlphaRLM variants also show substantially lower mean absolute
correlations than the fixed factor libraries and conventional automated
mining baselines across all three markets.

The upper tail provides a complementary view of redundancy.
Across every market and both model sizes, fewer than 0.6\% of AlphaRLM
factor pairs satisfy $|r|\geq0.8$.
By contrast, this proportion exceeds 22\% for Alpha360 in every market.
Relative to Alpha360, AlphaRLM reduces mean absolute pairwise correlation
by approximately 76--88\% across the three markets, while reducing the
share of highly correlated pairs by roughly 98\%.
Several LLM-based baselines, including QuantaAlpha and DeepSeek-R1, also
exhibit low pairwise correlations.
Overall, the results indicate that AlphaRLM's downstream gains are not
associated with linear redundancy in the final factor pools.

\subsection{Structural Diversity and Library Overlap}

\paragraph{Canonicalization protocol.}
We complement correlation-based redundancy with a syntax-level analysis
of the four final AlphaRLM factor pools.
Because CSI 300 and CSI 1000 share the same A-share factor pool, this pool
is counted only once.
Each expression is parsed into an abstract syntax tree (AST) and represented
at two levels.
Exact canonicalization preserves operators, market variables, and numerical
parameters while normalizing equivalent syntactic forms.
For parameter-normalized structural skeletons, temporal parameters are
replaced by a placeholder $W$ and other numerical constants by
$C$, while operator and variable identities are preserved.
Children of commutative operators such as \texttt{Add} and \texttt{Mul}
are sorted, whereas non-commutative operators retain their order.
We perform no algebraic simplification.

\paragraph{Within-pool structural diversity.}
As shown in Table~\ref{tab:factor_structural_diversity}, exact-expression
uniqueness ranges from 99.5\% to 99.9\%, indicating that direct duplicates
are rare in all pools.
After normalizing temporal windows and numerical parameters, structural
uniqueness remains between 80.7\% and 95.9\%.
The median skeleton-cluster size is one in every pool.
Repeated parameterizations are most prevalent for AlphaRLM-14B on A-shares,
where 28.1\% of factors belong to non-singleton clusters and the largest
cluster contains 22 expressions.
Nevertheless, even this pool retains 807 distinct structural skeletons
among 1,000 factors.
Overall, the final pools are not dominated by repeated parameterizations
of a small number of symbolic templates.

\begin{table*}[t]
\centering
\small
\caption{Sensitivity to the number of selected factors for AlphaRLM-8B.}
\label{tab:factor_count_sensitivity}
\begin{tabular}{lrrrrrr}
\toprule
Market & Factor count & AR (\%) & IC & RankIC & ICIR & MDD (\%) \\
\midrule
CSI 300 & 158 & 11.9469 & 0.023905 & 0.043301 & 0.147774 & -8.2226 \\
 & 360 & 12.6458 & 0.029587 & 0.046999 & 0.183973 & -7.1806 \\
 & 1,000 & 12.1375 & 0.025554 & 0.042442 & 0.158593 & -6.1504 \\
 & 1,500 & 11.1957 & 0.026342 & 0.043941 & 0.159020 & -7.4845 \\
 & 2,000 & 9.0978 & 0.025492 & 0.043189 & 0.155702 & -7.4613 \\
\midrule
CSI 1000 & 158 & 11.9513 & 0.032682 & 0.038445 & 0.256770 & -10.5107 \\
 & 360 & 13.4345 & 0.032707 & 0.037646 & 0.260574 & -10.1139 \\
 & 1,000 & 17.4520 & 0.030222 & 0.041396 & 0.239800 & -11.9812 \\
 & 1,500 & 14.7490 & 0.029095 & 0.041199 & 0.225420 & -15.1168 \\
 & 2,000 & 17.2427 & 0.029782 & 0.041190 & 0.232260 & -13.9748 \\
\midrule
S\&P 500 & 158 & 2.6711 & 0.002328 & 0.005226 & 0.018174 & -31.8369 \\
 & 360 & 5.2059 & 0.001511 & 0.004661 & 0.011447 & -24.6369 \\
 & 1,000 & 6.94 & 0.0021 & 0.0031 & 0.0158 & -14.84 \\
 & 1,500 & 2.2731 & 0.000327 & 0.006166 & 0.002262 & -25.5465 \\
 & 2,000 & -0.3283 & -0.000011 & 0.004123 & -0.000092 & -32.4691 \\
\bottomrule
\end{tabular}
\end{table*}

\paragraph{Overlap with standard factor libraries.}
We further apply the same canonicalization procedure to all Alpha158 and
Alpha360 expressions; their union contains 48 unique parameter-normalized
skeletons.
Exact expression overlap between AlphaRLM and these standard libraries is
at most 0.5\% across the four pools.
More importantly, only 0--0.337\% of AlphaRLM's unique structural
skeletons overlap with Alpha158 or Alpha360.
Even in the 8B S\&P 500 pool, which shows the highest structural overlap,
only three of 891 unique skeletons match the standard libraries, covering
eight of the 1,000 selected factors.
Together with the pairwise-correlation results above, these findings provide
complementary behavioral and structural evidence that AlphaRLM's final
factor pools are not dominated by highly correlated copies or parameterized
variants of standard library factors.
These statistics characterize syntactic structural diversity rather than
economic novelty and do not rule out reuse of common subexpressions.

\section{Sensitivity to the Number of Selected Factors}
\label{app:factor_count}

We examine how the number of selected factors affects downstream performance using AlphaRLM-8B.
Candidate factors are ranked by XGBoost feature importance using training-period data, and the default count of 1,000 is selected using training-period information.
For the sensitivity analysis, we retain the top $K$ factors from the same ranked candidate pool, with $K\in\{158,360,1000,1500,2000\}$.
All other settings follow the main experiment, including the downstream model, portfolio strategy, and five-seed aggregation.
Table~\ref{tab:factor_count_sensitivity} reports performance over the 2021--2024 test period for these training-ranked subsets.

The effect of pool size differs across markets.
On CSI 300, 360 factors achieve the highest AR (12.65\%), while 1,000 factors yield a lower AR of 12.14\% with the smallest drawdown magnitude (6.15\%).
On CSI 1000, 1,000 factors achieve the highest AR (17.45\%) and RankIC (0.041396); increasing the count to 1,500 or 2,000 produces lower AR and larger drawdowns.
On S\&P 500, the 1,000-factor setting achieves the highest AR (6.94\%) and smallest drawdown magnitude (14.84\%).
Expanding this pool to 1,500 and 2,000 reduces AR to 2.27\% and $-0.33\%$, respectively.

Predictive correlations and portfolio performance respond differently to pool size.
For example, S\&P 500 reaches its highest RankIC at 1,500 factors, despite substantially lower AR than at 1,000.
Across the evaluated settings, 1,000 factors offer a return--drawdown trade-off, while larger pools provide no consistent improvement.
These test-period results characterize the behavior of the training-selected configuration.

\section{Downstream Predictor Sensitivity}
\label{app:ridge_sensitivity}

\paragraph{Evaluation protocol.}
We examine whether AlphaRLM's downstream advantage depends on the nonlinear interaction capacity of XGBoost.
To isolate this effect, we retain the same selected factor pools, temporal splits, preprocessing, target definition, and portfolio construction as in the main experiments, while replacing only the downstream return predictor with Ridge regression.
The selected factor pools therefore remain fixed across the XGBoost and Ridge evaluations.
Ridge provides a deliberately restricted diagnostic: it combines factors linearly under $\ell_2$ regularization and cannot model the nonlinear feature interactions available to XGBoost.

\begin{table*}[t]
\centering
\small
\caption{
Backtest performance using Ridge regression as the downstream predictor.
The factor pools and evaluation protocol are held fixed relative to the main experiments, with only the return predictor replaced.
Within each dataset and metric, the best result is shown in \textbf{bold}, and the second-best result is \underline{underlined}.
}
\label{tab:ridge_results}

\renewcommand{\arraystretch}{1.18}
\setlength{\tabcolsep}{3pt}

\resizebox{\textwidth}{!}{
\begin{tabular}{lrrrrr|rrrrr|rrrrr}
\toprule
& \multicolumn{5}{c|}{\textbf{CSI 300}}
& \multicolumn{5}{c|}{\textbf{CSI 1000}}
& \multicolumn{5}{c}{\textbf{S\&P 500}} \\

Method
& AR & IC & RankIC & ICIR & MDD
& AR & IC & RankIC & ICIR & MDD
& AR & IC & RankIC & ICIR & MDD \\
\midrule

\rowcolor{gray!15}
\multicolumn{16}{c}{\textbf{\textit{Factor Libraries}}} \\

Alpha158
& 0.0624 & 0.0152 & 0.0333 & 0.0891 & -0.1357
& 0.0734 & 0.0010 & 0.0164 & 0.0074 & \textbf{-0.1500}
& -0.1358 & -0.0013 & 0.0017 & -0.0092 & -0.6318 \\

Alpha360
& 0.0175 & 0.0082 & 0.0209 & 0.0377 & -0.2501
& -0.0259 & 0.0065 & 0.0303 & 0.0489 & -0.2241
& -0.1864 & -0.0173 & -0.0191 & -0.1343 & -0.8209 \\

\midrule
\rowcolor{gray!15}
\multicolumn{16}{c}{\textbf{\textit{Automated Factor Mining}}} \\

GP-Learn
& 0.0150 & 0.0049 & 0.0068 & 0.0280 & -0.2109
& 0.0870 & 0.0166 & \underline{0.0365} & 0.1208 & -0.1761
& -0.0758 & -0.0042 & \underline{0.0040} & -0.0247 & -0.4063 \\

AlphaGen
& 0.0624 & 0.0129 & 0.0336 & 0.0724 & -0.2033
& \underline{0.0930} & \textbf{0.0194} & \textbf{0.0375} & \textbf{0.1562} & -0.1541
& -0.1174 & -0.0046 & 0.0002 & -0.0401 & -0.4898 \\

AlphaForge
& 0.0320 & 0.0061 & 0.0246 & 0.0316 & -0.1116
& 0.0787 & 0.0115 & 0.0281 & 0.0892 & \underline{-0.1531}
& -0.1244 & -0.0059 & 0.0007 & -0.0476 & -0.5348 \\

\midrule
\rowcolor{gray!15}
\multicolumn{16}{c}{\textbf{\textit{LLM-based Factor Mining}}} \\

DeepSeek-R1
& \underline{0.0763} & \textbf{0.0199} & \underline{0.0376} & \textbf{0.1304} & \underline{-0.0832}
& 0.0406 & 0.0060 & 0.0248 & 0.0418 & -0.2041
& -0.0662 & -0.0028 & 0.0003 & -0.0259 & -0.3740 \\

GLM-5.2
& -0.0143 & 0.0045 & 0.0254 & 0.0217 & -0.2096
& 0.0535 & 0.0037 & 0.0230 & 0.0252 & -0.2189
& -0.0697 & -0.0003 & 0.0037 & -0.0025 & \underline{-0.3053} \\

Fin-R1
& 0.0080 & 0.0062 & 0.0235 & 0.0294 & -0.2453
& 0.0860 & 0.0077 & 0.0289 & 0.0488 & -0.1909
& -0.1251 & -0.0019 & -0.0020 & -0.0163 & -0.5769 \\

FactorMiner
& 0.0260 & 0.0038 & 0.0176 & 0.0301 & -0.1039
& 0.0925 & 0.0167 & 0.0352 & \underline{0.1413} & -0.1682
& -0.0821 & \underline{0.0003} & 0.0011 & \underline{0.0035} & -0.4471 \\

QuantaAlpha
& 0.0484 & 0.0059 & 0.0239 & 0.0337 & -0.1187
& 0.0899 & \underline{0.0176} & 0.0341 & 0.1345 & -0.1825
& -0.1259 & \textbf{0.0010} & \textbf{0.0061} & \textbf{0.0085} & -0.5844 \\

\textbf{AlphaRLM (8B)}
& 0.0751 & 0.0142 & 0.0335 & 0.0821 & -0.0928
& 0.0856 & 0.0020 & 0.0220 & 0.0139 & -0.1708
& \textbf{-0.0390} & -0.0057 & 0.0010 & -0.0530 & \textbf{-0.3028} \\

\textbf{AlphaRLM (14B)}
& \textbf{0.0832} & \underline{0.0171} & \textbf{0.0380} & \underline{0.0955} & \textbf{-0.0764}
& \textbf{0.0958} & 0.0050 & 0.0242 & 0.0334 & -0.2000
& \underline{-0.0644} & -0.0001 & 0.0028 & -0.0009 & -0.3987 \\

\bottomrule
\end{tabular}
}
\end{table*}

\paragraph{Results.}
The Ridge results show a substantial decline in absolute portfolio performance relative to the nonlinear XGBoost pipeline, particularly on S\&P~500, where every method produces negative AR.
This is consistent with the deliberately restricted linear form of Ridge: the purpose of this experiment is not to replace the main nonlinear predictor, but to test whether AlphaRLM's relative advantage relies entirely on XGBoost-specific nonlinear interactions.

Despite this restriction, AlphaRLM retains the strongest AR in every market through one of its two model sizes.
AlphaRLM-14B ranks first on CSI~300 (8.32\%) and CSI~1000 (9.58\%), while AlphaRLM-8B ranks first on S\&P~500 ($-3.90\%$), with AlphaRLM-14B second ($-6.44\%$).
On CSI~300, the 14B model also achieves the highest RankIC and smallest drawdown, whereas predictive correlations are more mixed on CSI~1000 and S\&P~500.
Thus, Ridge does not reproduce the full absolute performance of the XGBoost pipeline, but the persistence of AlphaRLM's relative AR advantage indicates that nonlinear interactions learned by XGBoost are not the sole explanation for its downstream gains.

Together with the factor-level TMB analysis in Section~\ref{sec:factor_cases}%
, these results suggest that AlphaRLM's performance reflects both factor-level predictive structure and multivariate aggregation.

\raggedbottom
\section{Prompts}
\label{app:prompts}

\begin{promptbox}{Finance Relevance Classification Prompt}
Based on the title and abstract of the paper, please analyze whether this paper belongs to the field of quantitative finance.

\textbf{Title:} \{title\}

\textbf{Abstract:} \{abstract\}

If yes, output [YES]; otherwise, output [NO].
\end{promptbox}

The training prompt is split into two boxes for readability, preserving the original wording and all examples.
The editorial insertion marker indicates where the full syntax block appeared; Markdown formatting is rendered typographically.

\begin{promptbox}{Alpha Mining Prompt}
\par\textbf{Instruction}\par\nopagebreak
You are an expert in the financial domain, and your task is to uncover technical alpha factors. You can draw inspiration from the title and abstract of a paper.\par
Write a financial technical alpha factor. The alpha factor should have a certain level of uniqueness and informational advantage, and should not overlap too much with common simple alphas.\par

\par\textbf{Available Operators \& Factor Format}\par\nopagebreak
\textit{[Insert the full contents of the Qlib Factor Expression Syntax Rules box below.]}

\par\textbf{Paper Title \& Abstract}\par\nopagebreak
\{title\}\par
\{abstract\}\par
\par\textbf{Output}\par\nopagebreak
First, output your thought process and reasoning. In the final result, provide the factor expressions using:\par
\par\begingroup\footnotesize\ttfamily\raggedright
field = factor\_expression\par
\endgroup

\par\textbf{Attention}\par\nopagebreak
\par\noindent\textbullet\ You should ensure that only the operators and market data variables provided above are used.
\par\noindent\textbullet\ Time Series Operations can only be applied to time series and cannot be represented with subscript or superscript notation, such as using \textasciicircum{}2 to represent square.
The reasoning process and the answer are enclosed within \textless{}think\textgreater{} and \textless{}answer\textgreater{} tags, respectively, as follows:\par
\textless{}think\textgreater{} This is the reasoning process \textless{}/think\textgreater{}\par
\textless{}answer\textgreater{} This is the answer, including the formula list for fields, which should be included inside a Python code block \textless{}/answer\textgreater{}\par
\end{promptbox}

\begin{promptbox}{Qlib Factor Expression Syntax Rules}
\par\textbf{1. Basic Syntax Structure}\par\nopagebreak
Each formula consists of the following parts:\par
\par\noindent\textbullet\ Variables: For example, \texttt{\$close} represents the closing price, \texttt{\$open} represents the opening price.
\par\noindent\textbullet\ Operators: Such as \texttt{/} (division), \texttt{-} (subtraction).
\par\noindent\textbullet\ Functions: Such as \texttt{Ref} (reference historical data), \texttt{Mean} (average), \texttt{Std} (standard deviation).

\par\textbf{Example:}\par\nopagebreak
\par\begingroup\footnotesize\ttfamily\raggedright
field = \textquotedbl{}Ref(\$close, 5)/\$close\textquotedbl{}\par
\endgroup

Explanation:\par
\par\noindent\textbullet\ \texttt{Ref(\$close, \%d)}: Reference the closing price \texttt{d} days ago.
\par\noindent\textbullet\ \texttt{/ \$close}: Divide by the current closing price to calculate the relative change.

\par\textbf{2. Supported Variables}\par\nopagebreak

Qlib only supports the following market data variables.\par

\par\textbf{Price-related Variables:}\par\nopagebreak
\par\noindent\textbullet\ \texttt{\$close}: Closing price
\par\noindent\textbullet\ \texttt{\$open}: Opening price
\par\noindent\textbullet\ \texttt{\$high}: Highest price
\par\noindent\textbullet\ \texttt{\$low}: Lowest price
\par\noindent\textbullet\ \texttt{\$vwap}: Volume-weighted average price

\par\textbf{Volume-related Variables:}\par\nopagebreak
\par\noindent\textbullet\ \texttt{\$volume}: Trading volume

\par\textbf{3. Supported Functions}\par\nopagebreak

You can only use the functions below.\par

\par\textbf{Time Series Operations:}\par\nopagebreak
\par\noindent\textbullet\ \texttt{Ref}: Reference historical data
\par\noindent\textbullet\ \texttt{Max}: Rolling maximum
\par\noindent\textbullet\ \texttt{Min}: Rolling minimum
\par\noindent\textbullet\ \texttt{Sum}: Rolling sum
\par\noindent\textbullet\ \texttt{Mean}: Rolling average
\par\noindent\textbullet\ \texttt{Std}: Rolling standard deviation
\par\noindent\textbullet\ \texttt{Var}: Rolling variance
\par\noindent\textbullet\ \texttt{Skew}: Rolling skewness
\par\noindent\textbullet\ \texttt{Kurt}: Rolling kurtosis
\par\noindent\textbullet\ \texttt{Med}: Rolling median
\par\noindent\textbullet\ \texttt{Mad}: Rolling mean absolute deviation
\par\noindent\textbullet\ \texttt{Slope}: Rolling linear regression slope
\par\noindent\textbullet\ \texttt{Rsquare}: Rolling regression R-squared
\par\noindent\textbullet\ \texttt{Resi}: Rolling regression residual
\par\noindent\textbullet\ \texttt{Rank}: Rolling percentile rank
\par\noindent\textbullet\ \texttt{Quantile}: Rolling quantile
\par\noindent\textbullet\ \texttt{Count}: Rolling non-null value count
\par\noindent\textbullet\ \texttt{Delta}: Rolling difference
\par\noindent\textbullet\ \texttt{WMA}: Rolling weighted moving average
\par\noindent\textbullet\ \texttt{EMA}: Rolling exponential weighted average

\par\textbf{Element-wise Operations:}\par\nopagebreak
\par\noindent\textbullet\ \texttt{Abs}: Absolute value
\par\noindent\textbullet\ \texttt{Sign}: Sign function
\par\noindent\textbullet\ \texttt{Log}: Logarithm
\par\noindent\textbullet\ \texttt{Not}: Bitwise NOT
\par\noindent\textbullet\ \texttt{Mask}: Mask specific targets

\par\noindent\textbullet\ Binary Operators:
\par\noindent\textbullet\ Mathematical operations: \texttt{Power} (exponentiation), \texttt{Add} (addition), \texttt{Sub} (subtraction), \texttt{Mul} (multiplication), \texttt{Div} (division)
\par\noindent\textbullet\ Comparison operations: \texttt{Greater} (maximum), \texttt{Less} (minimum), \texttt{Gt} (greater than), \texttt{Ge} (greater than or equal), \texttt{Lt} (less than), \texttt{Le} (less than or equal), \texttt{Eq} (equal), \texttt{Ne} (not equal)
\par\noindent\textbullet\ Logical operations: \texttt{And} (bitwise AND), \texttt{Or} (bitwise OR)

\par\noindent\textbullet\ Feature Indexing:
\par\noindent\textbullet\ \texttt{IdxMax}: Rolling maximum index
\par\noindent\textbullet\ \texttt{IdxMin}: Rolling minimum index

\par\textbf{Composite Features:}\par\nopagebreak
\par\noindent\textbullet\ Ternary Operators:
\par\noindent\textbullet\ \texttt{If}: Conditional operation

\par\textbf{Time Operations:}\par\nopagebreak
\par\noindent\textbullet\ Only applicable to time-indexed data:
\par\noindent\textbullet\ \texttt{TResample}: Time resampling (e.g., aggregating data by specific frequency), parameters: 'B': Business day (excluding Saturday and Sunday), 'D': Day (daily), 'W': Week (default is weekly ending on Sunday), 'M': Month (month-end), 'SM': Semi-month (15th and month-end), 'Q': Quarter (quarter-end), 'A': Year (year-end); you can also create compound frequencies by combining time units, e.g., '2D': Every two days, '2W-MON': Every two weeks on Monday, '1Q-JAN': Every January (quarterly).

\par\textbf{Specific Rolling Operations:}\par\nopagebreak
\par\noindent\textbullet\ Binary Rolling Operators:
\par\noindent\textbullet\ \texttt{Corr}: Rolling correlation coefficient
\par\noindent\textbullet\ \texttt{Cov}: Rolling covariance

\par\textbf{PIT (Point-in-Time Operations):}\par\nopagebreak
\par\noindent\textbullet\ \texttt{PIT}: Point-specific operation
\par\noindent\textbullet\ \texttt{PRef}: Point reference operation

\par\textbf{4. Writing Standards}\par\nopagebreak
1. Operator writing order: For combinations of multiple operators, write in the following order:\par
\par\noindent\textbullet\ Parentheses: Used to control the priority of operations.
\par\noindent\textbullet\ Element-wise operators: \texttt{Abs}, \texttt{Sign}, \texttt{Log}, \texttt{Not}, etc.
\par\noindent\textbullet\ Binary operators: Such as \texttt{+}, \texttt{-}, \texttt{*}, \texttt{/}, \texttt{\textgreater{}}, \texttt{\textless{}}, \texttt{==}, etc.
\par\noindent\textbullet\ Ternary operators: \texttt{If}.
\par\noindent\textbullet\ Function calls: Such as \texttt{Rolling()}, \texttt{Ref()}, etc.

2. Use string formatting: For example, in \texttt{Ref(\$close, \%d)}, use \texttt{\%d} to format dates or other dynamically changing parts.\par

3. Naming conventions: Variable names usually start with \texttt{\$} (e.g., \texttt{\$close}), functions are usually in uppercase (e.g., \texttt{Rolling}), and operators are in lowercase (e.g., \texttt{add}).\par

\par\textbf{5. Example Formulas:}\par\nopagebreak

\par\textbf{Rolling (Rolling Window Operation)}\par\nopagebreak

N = 0, retrieve the first data; N \textgreater{} 0, retrieve data of N periods ago; N \textless{} 0, future data, so N should not be less than 0\par

\par\noindent\textbullet\ \texttt{Ref}: Reference data from a specific time in the past (e.g., the closing price 5 days ago).

\par\begingroup\footnotesize\ttfamily\raggedright
\# Reference the closing price 5 days ago\par
field = \textquotedbl{}Ref(\$close, 5)\textquotedbl{}\par
\endgroup

\par\noindent\textbullet\ \texttt{Max}: Calculate the rolling maximum.

\par\begingroup\footnotesize\ttfamily\raggedright
\# Calculate the maximum closing price of the past 10 days\par
field = \textquotedbl{}Max(\$close, 10)\textquotedbl{}\par
\endgroup

\par\noindent\textbullet\ \texttt{Min}: Calculate the rolling minimum.

\par\begingroup\footnotesize\ttfamily\raggedright
\# Calculate the minimum closing price of the past 10 days\par
field = \textquotedbl{}Min(\$close, 10)\textquotedbl{}\par
\endgroup

\par\noindent\textbullet\ \texttt{Sum}: Calculate the rolling sum.

\par\begingroup\footnotesize\ttfamily\raggedright
\# Calculate the sum of trading volume over the past 5 days\par
field = \textquotedbl{}Sum(\$volume, 5)\textquotedbl{}\par
\endgroup

\par\noindent\textbullet\ \texttt{Mean}: Calculate the rolling mean.

\par\begingroup\footnotesize\ttfamily\raggedright
\# Calculate the rolling mean of the past 10 days\par
field = \textquotedbl{}Mean(\$close, 10)\textquotedbl{}\par
\endgroup

\par\noindent\textbullet\ \texttt{Std}: Calculate the rolling standard deviation.

\par\begingroup\footnotesize\ttfamily\raggedright
\# Calculate the standard deviation of the past 10 days\par
field = \textquotedbl{}Std(\$close, 10)\textquotedbl{}\par
\endgroup

\par\noindent\textbullet\ \texttt{Var}: Calculate the rolling variance.

\par\begingroup\footnotesize\ttfamily\raggedright
\# Calculate the variance of the past 5 days\par
field = \textquotedbl{}Var(\$close, 5)\textquotedbl{}\par
\endgroup

\par\noindent\textbullet\ \texttt{Skew}: Calculate the rolling skewness.

\par\begingroup\footnotesize\ttfamily\raggedright
\# Calculate the rolling skewness of the past 10 days\par
field = \textquotedbl{}Skew(\$close, 10)\textquotedbl{}\par
\endgroup

\par\noindent\textbullet\ \texttt{Kurt}: Calculate the rolling kurtosis.

\par\begingroup\footnotesize\ttfamily\raggedright
\# Calculate the rolling kurtosis of the past 5 days\par
field = \textquotedbl{}Kurt(\$close, 5)\textquotedbl{}\par
\endgroup

\par\noindent\textbullet\ \texttt{Med}: Calculate the rolling median.

\par\begingroup\footnotesize\ttfamily\raggedright
\# Calculate the rolling median of the past 7 days\par
field = \textquotedbl{}Med(\$close, 7)\textquotedbl{}\par
\endgroup

\par\noindent\textbullet\ \texttt{Mad}: Calculate the rolling mean absolute deviation.

\par\begingroup\footnotesize\ttfamily\raggedright
\# Calculate the rolling mean absolute deviation of the past 10 days\par
field = \textquotedbl{}Mad(\$close, 10)\textquotedbl{}\par
\endgroup

\par\noindent\textbullet\ \texttt{Slope}: Calculate the rolling linear regression slope.

\par\begingroup\footnotesize\ttfamily\raggedright
\# Calculate the linear regression slope of the closing price over the past 30 days\par
field = \textquotedbl{}Slope(\$close, 30)\textquotedbl{}\par
\endgroup

\par\noindent\textbullet\ \texttt{Rsquare}: Calculate the rolling regression R-squared.

\par\begingroup\footnotesize\ttfamily\raggedright
\# Calculate the regression R-squared of the past 20 days\par
field = \textquotedbl{}Rsquare(\$close, 20)\textquotedbl{}\par
\endgroup

\par\noindent\textbullet\ \texttt{Resi}: Calculate the rolling regression residual.

\par\begingroup\footnotesize\ttfamily\raggedright
\# Calculate the regression residual of the past 20 days\par
field = \textquotedbl{}Resi(\$close, 20)\textquotedbl{}\par
\endgroup

\par\noindent\textbullet\ \texttt{Rank}: Calculate the rolling percentile rank.

\par\begingroup\footnotesize\ttfamily\raggedright
\# Calculate the rolling rank of the past 10 days\par
field = \textquotedbl{}Rank(\$close, 10)\textquotedbl{}\par
\endgroup

\par\noindent\textbullet\ \texttt{Quantile}: Calculate the rolling quantile.

\par\begingroup\footnotesize\ttfamily\raggedright
\# Calculate the rolling quantile of the past 5 days\par
field = \textquotedbl{}Quantile(\$close, 5, 0.5)\textquotedbl{}  \# Median\par
\endgroup

\par\noindent\textbullet\ \texttt{Count}: Calculate the rolling non-null value count.

\par\begingroup\footnotesize\ttfamily\raggedright
\# Calculate the non-null value count of the past 15 days\par
field = \textquotedbl{}Count(\$close, 15)\textquotedbl{}\par
\endgroup

\par\noindent\textbullet\ \texttt{Delta}: Calculate the rolling difference.

\par\begingroup\footnotesize\ttfamily\raggedright
\# Calculate the difference of the past 5 days\par
field = \textquotedbl{}Delta(\$close, 5)\textquotedbl{}\par
\endgroup

\par\noindent\textbullet\ \texttt{WMA}: Calculate the rolling weighted moving average.

\par\begingroup\footnotesize\ttfamily\raggedright
\# Calculate the weighted moving average of the past 5 days\par
field = \textquotedbl{}WMA(\$close, 5)\textquotedbl{}\par
\endgroup

\par\noindent\textbullet\ \texttt{EMA}: Calculate the rolling exponential weighted average.

\par\begingroup\footnotesize\ttfamily\raggedright
\# Calculate the exponential weighted average of the past 5 days\par
field = \textquotedbl{}EMA(\$close, 5)\textquotedbl{}\par
\endgroup

\par\textbf{Abs: Calculate the absolute value.}\par\nopagebreak

\par\begingroup\footnotesize\ttfamily\raggedright
\# Calculate the absolute value of the closing price\par
field = \textquotedbl{}Abs(\$close)\textquotedbl{}\par
\endgroup

\par\textbf{Sign: Calculate the sign function.}\par\nopagebreak

\par\begingroup\footnotesize\ttfamily\raggedright
\# Get the sign of the closing price\par
field = \textquotedbl{}Sign(\$close)\textquotedbl{}\par
\endgroup

\par\textbf{Log: Calculate the logarithm.}\par\nopagebreak

\par\begingroup\footnotesize\ttfamily\raggedright
\# Calculate the logarithm of the closing price\par
field = \textquotedbl{}Log(\$close)\textquotedbl{}\par
\endgroup

\par\textbf{Not: Calculate the bitwise NOT.}\par\nopagebreak

\par\begingroup\footnotesize\ttfamily\raggedright
\# NOT operation, for boolean values\par
field = \textquotedbl{}Not(\$close \textgreater{} 100)\textquotedbl{}\par
\endgroup

\par\textbf{Mask: Mask specific targets.}\par\nopagebreak

\par\begingroup\footnotesize\ttfamily\raggedright
\# Mask specific targets (e.g., calculate indicators for specific stocks)\par
field = \textquotedbl{}Mask(\$close, \textquotesingle{}AAPL\textquotesingle{})\textquotedbl{}\par
\endgroup

\par\textbf{Power: Calculate the power.}\par\nopagebreak

\par\begingroup\footnotesize\ttfamily\raggedright
\# Calculate the square of the closing price\par
field = \textquotedbl{}Power(\$close, 2)\textquotedbl{}\par
\endgroup

\par\textbf{Add: Calculate addition.}\par\nopagebreak

\par\begingroup\footnotesize\ttfamily\raggedright
\# Calculate the sum of the closing price and the opening price\par
field = \textquotedbl{}Add(\$close, \$open)\textquotedbl{}\par
\endgroup

\par\textbf{Sub: Calculate subtraction.}\par\nopagebreak

\par\begingroup\footnotesize\ttfamily\raggedright
\# Calculate the closing price minus the opening price\par
field = \textquotedbl{}Sub(\$close, \$open)\textquotedbl{}\par
\endgroup

\par\textbf{Mul: Calculate multiplication.}\par\nopagebreak

\par\begingroup\footnotesize\ttfamily\raggedright
\# Calculate the product of the closing price and the trading volume\par
field = \textquotedbl{}Mul(\$close, \$volume)\textquotedbl{}\par
\endgroup

\par\textbf{Div: Calculate division.}\par\nopagebreak

\par\begingroup\footnotesize\ttfamily\raggedright
\# Calculate the ratio of the closing price to the opening price\par
field = \textquotedbl{}Div(\$close, \$open)\textquotedbl{}\par
\endgroup

\par\textbf{Greater: Calculate the maximum.}\par\nopagebreak

\par\begingroup\footnotesize\ttfamily\raggedright
\# Calculate the maximum of the closing price and the opening price\par
field = \textquotedbl{}Greater(\$close, \$open)\textquotedbl{}\par
\endgroup

\par\textbf{Less: Calculate the minimum.}\par\nopagebreak

\par\begingroup\footnotesize\ttfamily\raggedright
\# Calculate the minimum of the closing price and the opening price\par
field = \textquotedbl{}Less(\$close, \$open)\textquotedbl{}\par
\endgroup

\par\textbf{Gt: Determine the greater than relationship.}\par\nopagebreak

\par\begingroup\footnotesize\ttfamily\raggedright
\# Determine if the closing price is greater than the opening price\par
field = \textquotedbl{}Gt(\$close, \$open)\textquotedbl{}\par
\endgroup

\par\textbf{Ge: Determine the greater than or equal relationship.}\par\nopagebreak

\par\begingroup\footnotesize\ttfamily\raggedright
\# Determine if the closing price is greater than or equal to the opening price\par
field = \textquotedbl{}Ge(\$close, \$open)\textquotedbl{}\par
\endgroup

\par\textbf{Lt: Determine the less than relationship.}\par\nopagebreak

\par\begingroup\footnotesize\ttfamily\raggedright
\# Determine if the closing price is less than the opening price\par
field = \textquotedbl{}Lt(\$close, \$open)\textquotedbl{}\par
\endgroup

\par\textbf{Le: Determine the less than or equal relationship.}\par\nopagebreak

\par\begingroup\footnotesize\ttfamily\raggedright
\# Determine if the closing price is less than or equal to the opening price\par
field = \textquotedbl{}Le(\$close, \$open)\textquotedbl{}\par
\endgroup

\par\textbf{Eq: Determine the equal relationship.}\par\nopagebreak

\par\begingroup\footnotesize\ttfamily\raggedright
\# Determine if the closing price is equal to the opening price\par
field = \textquotedbl{}Eq(\$close, \$open)\textquotedbl{}\par
\endgroup

\par\textbf{Ne: Determine the not equal relationship.}\par\nopagebreak

\par\begingroup\footnotesize\ttfamily\raggedright
\# Determine if the closing price is not equal to the opening price\par
field = \textquotedbl{}Ne(\$close, \$open)\textquotedbl{}\par
\endgroup

\par\textbf{And: Bitwise AND.}\par\nopagebreak

\par\begingroup\footnotesize\ttfamily\raggedright
\# Determine if the closing price is greater than 100 and the trading volume is greater than 5000\par
field = \textquotedbl{}And(\$close \textgreater{} 100, \$volume \textgreater{} 5000)\textquotedbl{}\par
\endgroup

\par\textbf{Or: Bitwise OR.}\par\nopagebreak

\par\begingroup\footnotesize\ttfamily\raggedright
\# Determine if the closing price is greater than 100 or the trading volume is greater than 5000\par
field = \textquotedbl{}Or(\$close \textgreater{} 100, \$volume \textgreater{} 5000)\textquotedbl{}\par
\endgroup

\par\textbf{If: Conditional operator.}\par\nopagebreak

\par\begingroup\footnotesize\ttfamily\raggedright
\# If the closing price is greater than the opening price, return 1; otherwise, return 0\par
field = \textquotedbl{}If(\$close \textgreater{} \$open, 1, 0)\textquotedbl{}\par
\endgroup

\par\textbf{TResample: Time resampling (e.g., aggregating data by specific frequency).}\par\nopagebreak

\par\begingroup\footnotesize\ttfamily\raggedright
\# Aggregate the closing price by week, summing\par
field = \textquotedbl{}TResample(\$close, \textquotesingle{}W\textquotesingle{}, \textquotesingle{}sum\textquotesingle{})\textquotedbl{}\par
\endgroup

\par\textbf{Corr: Rolling correlation coefficient.}\par\nopagebreak

\par\begingroup\footnotesize\ttfamily\raggedright
\# Calculate the rolling correlation coefficient between the closing price and the opening price (past 10 days)\par
field = \textquotedbl{}Corr(\$close, \$open, 10)\textquotedbl{}\par
\endgroup

\par\textbf{Cov: Rolling covariance.}\par\nopagebreak

\par\begingroup\footnotesize\ttfamily\raggedright
\# Calculate the rolling covariance between the closing price and the opening price (past 10 days)\par
field = \textquotedbl{}Cov(\$close, \$open, 10)\textquotedbl{}\par
\endgroup

\end{promptbox}

\end{document}